\documentclass[journal]{IEEEtran}
\usepackage{amsfonts}
\usepackage{pseudocode}
\usepackage{algorithmicx}
\usepackage{algorithm,algpseudocode}

\usepackage{xcolor}

\usepackage{pgfplots}
\pgfplotsset{compat=newest}
\usetikzlibrary{plotmarks}
\usetikzlibrary{arrows.meta}
\usepgfplotslibrary{patchplots}
\usepackage{grffile}
\usepackage{amsmath}

\usepackage{amsmath,amssymb,amsthm}
\usepackage{graphicx}
\usepackage{subfigure}
\usepackage{color}
\usepackage{graphics}
\usepackage{verbatim,bm}
\usepackage{setspace}
\usepackage{enumitem}
\usepackage{bbm}

\usepackage{booktabs}

\title{DeepPool:  Distributed Model-free Algorithm for Ride-sharing using Deep Reinforcement Learning}
\author{Abubakr Alabbasi, Arnob Ghosh, and Vaneet Aggarwal\thanks{The authors are with Purdue University, West Lafayette, IN 47907, email:\{aalabbas,ghosh39,vaneet\}@purdue.edu }}

\begin{document}
\maketitle
%

%





\begin{abstract}
The success of modern ride-sharing platforms crucially depends on the profit of the ride-sharing fleet operating companies, and how efficiently the resources are managed. Further, ride-sharing allows sharing costs and, hence, reduces the congestion and emission by making better use of vehicle capacities. In this work, we develop a distributed model-free, DeepPool, that uses deep Q-network (DQN) techniques to learn optimal dispatch policies by interacting with the environment. Further, DeepPool
efficiently incorporates travel demand statistics and deep learning models to manage dispatching vehicles for improved ride sharing services.
Using real-world dataset of taxi trip records in New York City, DeepPool performs better than other strategies\textcolor{black}{, proposed in the literature,} that do not consider ride sharing or do not dispatch the vehicles to regions where the  future demand is anticipated. Finally, DeepPool can adapt rapidly to dynamic environments since it is implemented in a distributed manner in which each vehicle solves its own DQN individually without coordination.  

%
%
%
%
%
%
%
\end{abstract}


\section{Introduction}

Ride-sharing \textcolor{black}{(joint travel of two or more persons in a single vehicle)} has a background dating back to the times during World War 2 when there was a dire shortage of gas which resulted in shared rides, and later, in the 1970s during both the oil and energy crises \cite{hahn2017ridesharing}. Recently, it has  generated  interest because of the proliferation of ride-sharing apps such as Uber \cite{uberPool}  and Lyft  \cite{LyftLine}. It is estimated that sharing economy will increase from \$14 billion in 2014 to \$335 billion by 2025 \cite{yaraghi2017current}. Uber had 40 million active rides monthly worldwide, and has over 77\% of the US ride-hailing market share in 2016 \cite{kokalitcheva2016uber}.  A large portion of this revenue comes from ride-sharing with one vehicle serves multiple ride requests simultaneously. 

Ride-sharing or carpooling has a lot of potential to improve the socio-economic benefit. For example, ride-sharing aims to bring travelers together. It reduces the number of cars, and thus, it can alleviate the \textcolor{black}{congestion, traffic and and collisions \cite{li2013collisions,shaheen2018benefits}}. By dispatching the travelers together, it reduces the overall distance traversed compared to the scenario where only one car serves each traveler. Given the increase in autonomous vehicles \cite{litman2017autonomous} and electric vehicles, it becomes imperative to promote the ride-sharing in order to reduce the congestion and the energy usage \textcolor{black}{\cite{teubner2015economics}}. Thus, an optimal online dispatch algorithm is required to dispatch the cars to minimize the total distance traversed by the cars and the traveling time and waiting time for users. 

 Taxi-dispatch over a large city for carpooling is inherently challenging. It requires an instantaneous decision making for thousands of drivers to serve the users' requests over an uncertain future demand. A potential partially empty or empty car can be dispatched to serve future demands at certain location. The decision also depends on the users' behavior residing within the car.  For example, the total travel time of the users should not be large if a car takes a detour to pick up and drop off another customer. Hence, in a carpooling, a vehicle  also needs to decide whether it remains active to serve a new passenger. Thus, solving these challenges simultaneously is inherently difficult; the dispatch solution for each car is itself may be time consuming, exacerbating the uncertainty challenge of the optimization over a rapidly changing users' demand. Each car can be dispatched only if the car is not entirely full, and \textcolor{black}{the travel time of a user (if, any) within the car should be minimized}. Thus, the dispatch decision should depend on the future potential trip requests which are uncertain in nature. Yet, realistic models are needed to assess the trade-offs between potential conflicting interests, minimizing the unserved requests, waiting time, total travel times of users, and the vehicles' total trip distance.  
 
 Effective and efficient optimization technology that matches drivers and riders in
real-time is one of the necessary components for a successful dynamic ride-share system \cite{zhu2016public,agatz2012optimization,greenwood2017show}. The above approaches require a pre-specific models for evolution of the demand, the user's utilities for waiting times and travel times. It also requires a specific cost functions governing the cars' total trip distance which is difficult to estimate in practice. However, in a dynamic environment such as fleet management for carpooling the trip time and the actual routes the vehicles should take must be updated continuously based on the historic information. Thus, the objective function evolves in a complicated manner based on the action and the uncertainties involved in the number of available vehicles, the passengers' destination, the vehicles' locations, the demand, and the time a user already has spent in the car. Thus, a model based approach is unable to consider all the intricacies involved in the optimization problem. Further, a distributed algorithm is preferable where each vehicle can take its own decision without coordinating with the others. This paper aims to consider a model-free approach for ride-sharing with carpooling, which can adapt to the changing distributions of customer arrivals, the changing distributions of the active cars, the vehicles' locations, and user's valuations for the total travel time. The tools from deep reinforcement learning \cite{mnih2015human} are used for this dynamic modeling, where the transition probability is dynamically learnt using the deep neural network, based on which the action is chosen with the Q-learning strategy.

This paper proposes a model-free method for assigning vehicles to customers. We develop a distributed  optimized framework for vehicle dispatching (so-called DeepPool) that uses deep neural networks and reinforcement learning to learn the optimal policies for each vehicle individually by interacting with the external environment. Unlike existing model-based approaches, we do not need to accurately model the system (so we call it model-free), rather, we use reinforcement learning technique in which each vehicle solves its own deep Q-network (DQN) problem in a distributed manner without coordination with all other vehicles, and thus results in a significant reduction in the complexity. We note that this is the first work, to the best of our knowledge, that casts the ride-sharing problem (with pooling) to a reinforcement learning problem. Our goal is to optimally dispatch the vehicles to different locations in a service area as to minimize a set of performance metrics such as customers waiting time, extra travel time due to participating in ride-sharing and idle driving costs. We summarize our contributions as follows:
\begin{itemize}
\item We formulate an optimization problem where a dispatcher seeks to minimize four objectives-- i) the supply and demand mis-match, ii) the waiting time of the customers as well as the total time vehicles need to traverse to serve potential future demands, iii) the additional time spent by the users because of the car pooling, (iv) the number of used vehicles/resources and thus minimizing the fuel consumption and traffic congestion. 

	\item We propose a distributed algorithm, DeepPool, a model-free based approach, for dispatching the vehicles in large-scale systems. Each vehicle decides its  action by learning the impact of its action on the reward using a Q-learning method without co-ordinating with  other vehicles.    
	
	\item We use real-world dataset of taxi trip records in New York City \cite{manh_dataSet} (15 million taxi trip dataset) to design a modular large-scale realistic simulator to simulate the ride-sharing system. DeepPool ensures scalability by adopting a modular architecture and using deep neural networks for training such that each individual vehicle learns its own optimal decision independently.
%
	
	\item Evaluation results show the superiority of our approach (DeepPool) as compared to the state-of-the art algorithms, i.e., \cite{oda2018movi} and some considered baselines.  Further, \textcolor{black}{ empirical results show} that our approach can significantly reduce the number of operating vehicles (by at least $500$ vehicles) for the same request acceptance rate and the same wait time of the passengers. Thus, our algorithm has the capability to reduce the traffic congestion.
	
\end{itemize}

%

The rest of this paper is organized as follows. Section II describes some related work for this paper. 
In Section III, we state the problem and present an example for ride-sharing. Further, the performance metrics used in this paper are briefly explained.  
Section IV explains the proposed framework and the main components in the system design. In addition, DeepPool and DQN are also presented in Section V. Section VI gives the evaluation results and highlight the main findings out of this work, followed by a discussion on the advantage of DeepPool in Section VI. Section VII concludes the paper.

\section{Related Work}
\textcolor{black}{Ridesharing poses several challenges that have been extensively studied
in a number of works in the Artificial Intelligence literature. However, the majority of these works} focus on model-based approaches, e.g., \cite{zheng2017online,zhang2016control,ma2015real,kleiner2011mechanism,gopalakrishnan2016costs,bei2018algorithms} and references therein, which models the pickup request locations, travel time, destination and then try to come up with dispatching policy that would improve the performance. However, this type of work lacks adaptability and becomes very challenging when the number of states is very large, which is the case in our problem. Further, some algorithms for ride-sharing have been also studied, see for instance \cite{kleiner2011mechanism,gopalakrishnan2016costs}, where the goal is to design incentive compatible approaches to provide fair and efficient solutions. \textcolor{black}{In \cite{ma2015real}, a model-based mobile real-time taxi-sharing system is presented. The authors have scheduled proper taxis to customers subject to capacity, monetary cost, and time. In \cite{bistaffa2017cooperative}, a cooperative-game theoretic approach is proposed to tackle the ride-sharing problem. The problem is formulated as a graph-constrained coalition formation (GCCF) and then used CFSS algorithm to solve it.  The ridesharing problem is also cast under the pickup and delivery problem (PDP), e.g., \cite{dumitrescu2010traveling,furuhata2013ridesharing}. Authors in \cite{lu2015optimization} have optimized the number of miles driven by
drivers by pooling riders; ride requests are generated uniformly over square blocks. Based on analyzing
real-world data from a ride-sharing service, the authors in \cite{jauhri2017space} have modeled the ride requests and their variations over location and time using a graph. The spatial and temporal variability of ride requests and the potentials for ride pooling are captured. A matching algorithm is proposed in \cite{de2015comewithme} where the
utility of the user’s desired activity at the
destination is exploited for vehicles-riders matching. In \cite{zhang2017taxi}, the advantages and effects of taxi carpooling mode are analyzed by simulation. Results show that carpooling can decrease average payment of
passenger, increase income of driver, and improve passenger capacity.  
In \cite{ta2018efficient} vehicles and ride requests are treated as nodes of a bigraph, then
a maximum weighted matching is computed to manage the vehicles and requests. 
By utilizing 
the extensive cellular coverage and the high accuracy in position/velocity measurements provided by GPS devices, a traffic monitoring system based on GPS-enabled smartphones is proposed in \cite{herrera2010evaluation}. In \cite{psaraftis1980dynamic}
Psaraftis developed a dynamic-programming exact algorithm for both the static and dynamic versions of the Dial-a-Ride Problem (DARP) with a single vehicle, similar work can be found also in \cite{ho2018survey}.\\
Recent works consider different approaches for dispatching vehicles and allocating transportation resources in modern cities. Zhang et al. \cite{zhang2016control} have proposed an optimal method for autonomous vehicles, which considers both global service fairness and future costs. However, 
idle driving distance and real-time GPS information are not considered. in work In \cite{yang2004real,bertsimas2019online}, scheduling temporal-based methods are developed to reduce costs of idle cruising, however, the proposed algorithm does not utilize the real-time location information. Minimizing the total customer waiting time by concurrently dispatching multiple taxis and allowing taxis to exchange their booking assignments is investigated in \cite{seow2010collaborative}.  Lee et al. have proposed a dispatch system so as a taxi is dispatched through the shortest-time path by utilizing the real-time traffic conditions \cite{lee2004taxi}. A similar approach is proposed in \cite{xu2018large} where an algorithm for vehicle dispatching in a large-scale on-demand ride-hailing platforms is proposed. In \cite{powell2011towards,qu2014cost,huang2012detecting,zhang2014carpooling}, routing recommendations are provided to maximize drivers profits. While previous works provide promising results, authors only considered the current ride requests and available vehicles, i.e., does not neither utilize the statistics of travel demand nor incorporate deep learning techniques to better use of vehicles. Our design  exploits these two features to manage dispatching vehicles and thus improving ridesharing services.\\
Several previous studies show that it is possible to learn from past taxi data and thus organizing the taxi fleet and minimizing the wait-time for passengers and drivers
\cite{miao2016taxi,zhang2016framework,zhao2016predicting,zhang2017taxi,oda2018movi,oda2018distributed,kheterpal2018flow}. Nishant et. al in \cite{kheterpal2018flow}, have developed an open source based on reinforcement learning for traffic Control, aiming to improve traffic flow in a wide variety of traffic scenarios. A new set of benchmarks for mixed-autonomy traffic that expose several key
aspects of traffic control is discussed in \cite{vinitsky2018benchmarks,kreidieh2018dissipating}. However, our problem is different in settings, objective and formulations. In \cite{nam2017model}, authors
use Deep learning to solve traffic problems, such as travel
mode choice predication.
}
In \cite{oda2018movi}, authors used DQN based approach for dynamic fleet management and show that distributed DQN based approaches can outperform the centralized ones. However, this approach is only for dispatching the vehicles and does not consider ride-sharing. Hence, a vehicle can serve only one ride request. In this work, we consider a ride-sharing scenario and introduce different control knobs to improve ride-sharing services. 
%
%
%
%
%
%
%

\begin{figure}
	\begin{center}
		\includegraphics[trim=0.1in 2.1502in 0.2in .0in, clip,width=0.48\textwidth]{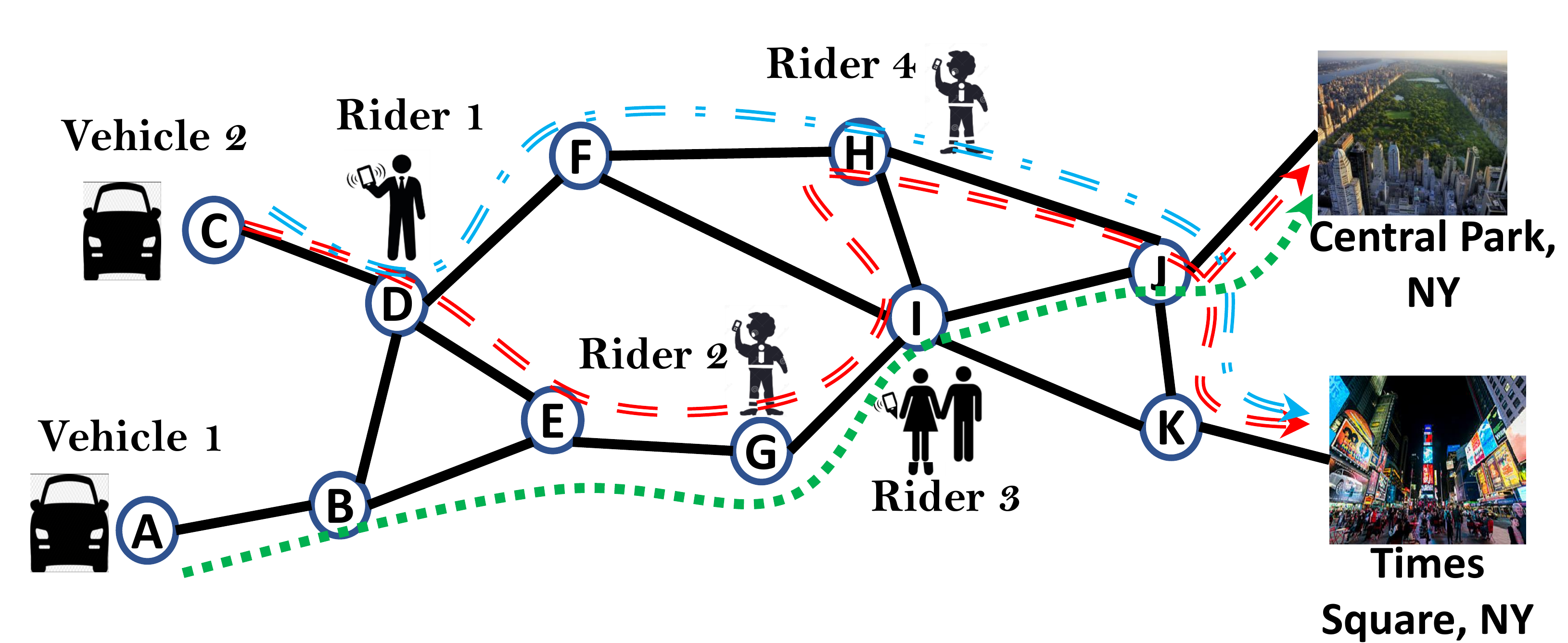}
			\vspace{-.1in}
		\caption{A schematic to illustrate the ride-sharing routing in a region graph, \textcolor{black}{ consisting of 11 regions, $A$ to $K$.} There are four ride requests and two vehicles. The locations of both customers and vehicles are shown in the figure above. Two different possible scenarios to serve the ride requests are shown in the figure and depicted by the dashed-red, dotted-green, and dashed-dotted blue lines. The destination of Rider 1 and Rider 4 is the Central Park, NY, while the destination of Rider 2 and Rider 3 is Times Square, NY.  	
		}
		\label{rideSharing_Example}
		\vspace{-.2in}
	\end{center}
\end{figure} 

\section{Problem Statement} \label{sec_probForm}

We consider a ride-sharing system such as UberPool \cite{uberPool} and Lyft \cite{LyftLine}, where multiple customers with similar itineraries are scheduled to share a vehicle. Each passenger requests for a ride  using an application on a mobile/desktop client. The vehicles need to be dispatched to pick up the passengers. Each vehicle's location, availability status, number of vacant seats, and passengers pickup requests can be known fairly accurately in real time, thanks to  mobile internet technologies.   
%

We assume that the area is divided into several non-overlapping zones. The zone is assumed to be small (e.g., 1-sq mile grid). A zone comprises of several locations. We take the decision which zones the vehicles should go. If the number of zones is large, naturally, the number of decision variables will increase. Each zone has 2 dimensions corresponding to longitude and latitude. Note that ride hail services such as Uber, Lyft always provide the heat map the zone where the demand is high. 

{ \bf Ride-sharing Scenario:} \textcolor{black}{Figure \ref{rideSharing_Example} presents an example depicting a real life scenario}. Consider the scenario shown in Figure \ref{rideSharing_Example}, where Rider 1 and Rider 4 want to go to the \textcolor{black}{Times Squares, NY}, while
Rider 2, and Rider 3 want to go to the \textcolor{black}{Central Park, NY.}
The locations of the four customers are depicted in the figure. Also, there are two vehicles located at node (zone) $A$ and node $C$\footnote{We use node, zone, and region interchangeably. However, a node also can refer to a certain location inside a region/zone.}. Without loss of generality, we assume the capacity of the vehicle ($2$) at location $C$ is $5$ passengers while that at location $A$ ($1$) is limited to $4$ passengers only. Given the two vehicles, there is more than one way to serve the requests. Two different possibilities are shown in the figure, assuming ride-sharing is possible. The two of the many possibilities are: (i) serving all the ride requests using only the vehicle $2$, depicted by the dashed-red line in the figure, (ii) serving Rider 1 and Rider 4 using the vehicle $2$ (see the dashed-dotted blue line), while Rider $2$ and Rider $3$ are served using the vehicle $1$\textcolor{black}{(see the dotted-green line)}. 
%
%
If ride-sharing is not allowed, only two ride requests among the four can get served at a given time. For example, vehicle $2$ needs to pick  up and dispatch rider $1$ first, then, it can pick up rider $4$. Though rider $4$'s location is inside the route of the rider $1$s source and destination. Hence,  
%
using ride-sharing, less resources \textcolor{black}{(only one vehicle, at location $C$, out of the two)} are used and consequently \textcolor{black}{less fuel and emission are consumed \cite{shaheen2018benefits}}. 
Further, the payment cost per rider should be lower because of sharing the cost among all riders. In addition, ride-sharing reduces traffic congestion and vehicles emission by better utilizing the vehicles seats. Thus, ride-sharing will bring benefits to the driver, riders, and society. 

Having explained the benefits of ride-sharing, we aim to design an optimal (or, near optimal) ride-sharing system that exploits all of these features and thus improve the ride-sharing service. 

We first introduce some notations which we use throughout this paper. We use $i\in\{1,2,3,\ldots,M\}$ as the zone index. Also, we assume the number of vehicles is $N$. We seek to optimize the system over $T$ time slots ahead. Each of the slots has length $\Delta t$, i.e., we seek to take a decision at each time slot  $\tau=t_0,t_0+1,t_0+2,\cdots,t_0+T$, where $t_0$ is the current time slot.  Let the number of available vehicles at region $i$ at time slot $t$ be ${v}_{t,i}$. A vehicle is marked {\em available} if at least one of its seats is vacant and the vehicle decides to pick up new customers, otherwise it is unavailable. Note that a vehicle is unavailable if either the vehicle is full or it decides against new passengers since it wants to drop off the existing passengers. 
We can only dispatch those vehicles which are available. Here, each vehicle has a limited seating, with vehicle $v$ having a \textcolor{black}{maximum} capacity of $C_v$ passengers. Similarly, we assume the number of requests at zone $i$ at time $t$ is ${d}_{t,i}$. \textcolor{black}{Let $d_{t,\widetilde{t},i}$ be the number of vehicles that are busy (not available) at time $t$ but will drop-off one (or more) of its occupants at zone $i$ and then be available at time $\widetilde{t}$. Given a set of dispatch actions/decisions, $d_{t,\widetilde{t},i}$ can be estimated}.

We now describe the state variables. We use $\boldsymbol{X}_t=\{\boldsymbol{x}_{t,1}, \boldsymbol{x}_{t,2},\ldots,\boldsymbol{x}_{t,N}\}$ to denote the vehicles status at time $t$. $\boldsymbol{x}_{t,k}$ is a vector which consists of the current location (the zone index) \textcolor{black}{of vehicle $k$}, available vacant seats (idle/occupied), the time at which a passenger (or more) has been/will be picked up, and destination of each passenger (the zone index). Using this information, we can also predict the time slot at which the unavailable vehicle $v$ will be available,  \textcolor{black}{$d_{t,\widetilde{t},i}$}, given the dispatch decisions. Thus, given a set of dispatch actions at time $t$, we can predict the number of vehicles in each zone for $T$ time slots ahead, denoted by $\boldsymbol{V}_{t:t+T}$.
 Though we cannot know the future demands exactly, customers usually show daily and weekly behaviors and hence their statistics can be learned to further improve the dispatching policies \cite{wyld2005my}. 
We use $\boldsymbol{D}_{t:T}=\left( \overline{\boldsymbol{d}}_{t},\ldots,\overline{\boldsymbol{d}}_{t+T} \right) $ to denote the predicted future demand from time $t_0$ to time $t+T$ at each zone. Combining this data, the environment state at time $t$ can be written as 
$\boldsymbol{s}_t=\left(\boldsymbol{X}_t, \boldsymbol{V}_{t:t+T}, \boldsymbol{D}_{t:t+T} \right)$. Note that when a passenger's request is accepted, we will append the user's expected pick-up time, the source, and destination.

\textcolor{black}{We now illustrate the state variables using a figure. Again consider the scenario depicted in Figure \ref{rideSharing_Example}. Consider the dispatch decision (ii), i.e., serving Rider 1 and Rider 4 using the vehicle $2$ (see the dashed-dotted blue line), while Rider $2$ and Rider $3$ are served using the vehicle $1$ (see the dotted-green line). Initially, $\boldsymbol{x}_{0,1}=\{{\bf A}, 4\}$ and $\boldsymbol{x}_{0,2}=\{{\bf C}, 5\}$. Suppose that the time(s) at which passenger $i$ will be picked up is $t_i$, thus, $\boldsymbol{x}_{t_2,1}=\{{\bf G}, 3,t_2,K\}$,  $\boldsymbol{x}_{t_1,2}=\{{\bf D}, 4,\boldsymbol{t}_1,J\}$. The entries in both vectors are, respectively, the current location of the vehicle, available vacant (idle) seats, the time(s) at which passenger $i$ has been picked up is $t_i$, and destination zones of each passenger. Similarly, we can write
	  $v_{t_2,D}=1$,  $v_{t_2,G}=1$, $d_{t_1,D}=1$,  $d_{t_4,H}=1$, $d_{t_2,G}=1$, $d_{t_3,I}=1$, $d_{t,E}=d_{t,F}=0$, etc. Note that both $v_{t,i}$ and $d_{t,i}$ vary based on the available vehicles and ride requests, respectively. In our model, those variables change in real time according to the environment variations and demand/supply dynamics. However, our framework keeps tracking of these rapid changes and seeks to make the demand, $\boldsymbol{d}_t, \forall t$  and supply $\boldsymbol{v}_t, \forall t$ close enough (i.e., mismatch between them is zero).  
}

\textcolor{black}{
	\begin{figure}[hbtp]
		\includegraphics[trim=0.0in 0.05in 0.250in 0.0in, clip,width=0.40\textwidth]{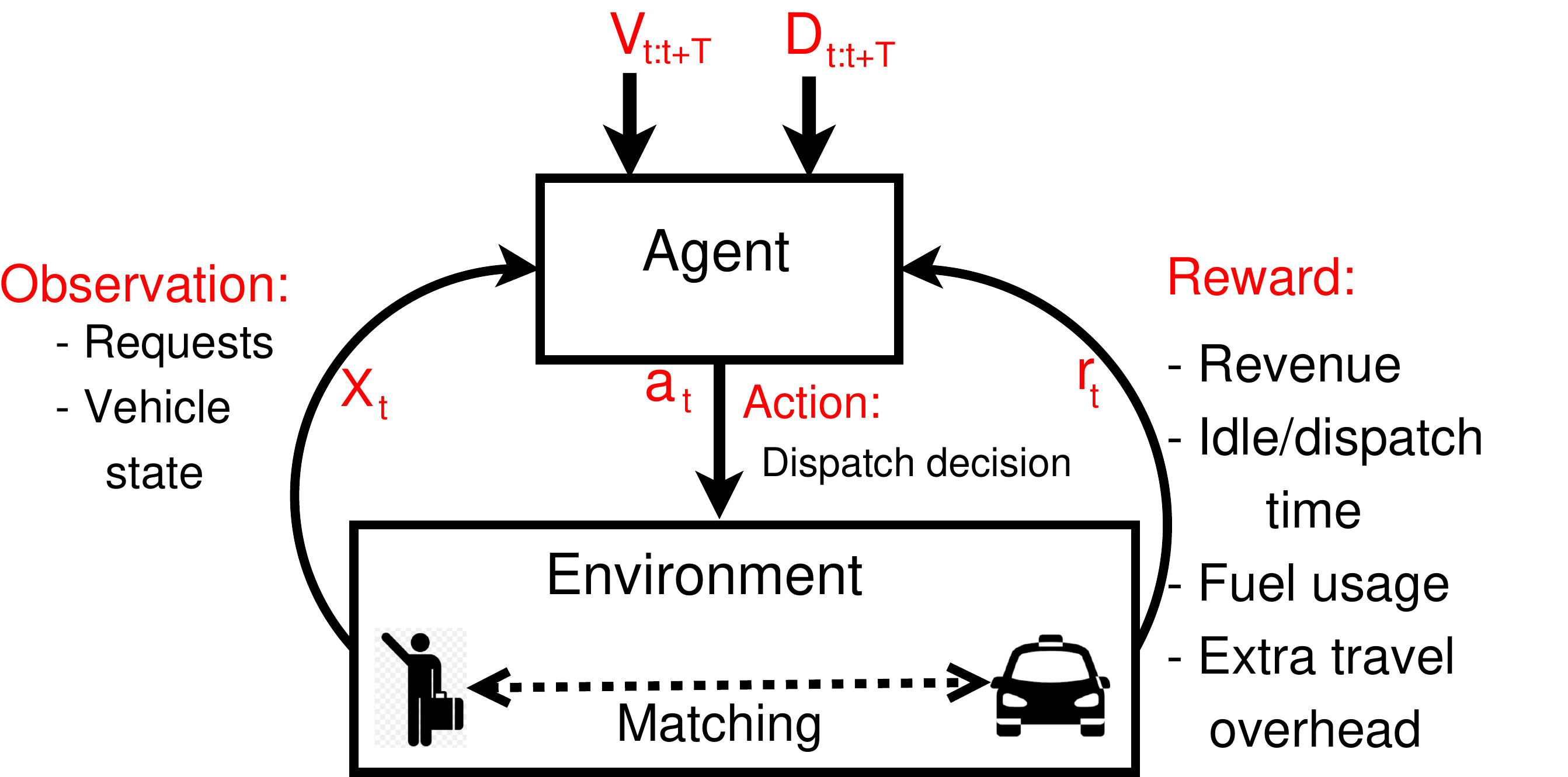}
		\caption{\textcolor{black}{The agent and the environment interactions. Based on the state observation $\boldsymbol{s}_t$ and the reward $\boldsymbol{r}_t$, an action $\boldsymbol{a}_t$ is taken accordingly. }}
		\label{RL_agent}
		\vspace{-0.5cm}
	\end{figure}
}
%
\textbf{Objectives}:
Our objective is to efficiently dispatch the available fleet of vehicles to different locations in a given area in order to achieve the following goals: (i) satisfy the demand (or equivalently minimize the demand-supply mismatch), (ii) minimize the waiting time of the customers (time elapsed between the ride request and the pick up), as well as the dispatch time which is the time a vehicle takes to move to another zone to pick up new customer (may be, future customers), (iii) the extra travel time due to participating in ride-sharing, and (iv) the number of used vehicles/resources as to minimize the fuel consumption, congestion, and to better utilize the vehicles seats. 
%
We will explain these metrics mathematically in detail later (Section~\ref{sec:objective}). These metrics can be summed with different weights to give a combined objective. Further, the weights of these different components can be varied to reflect the importance of each one to the customers and fleet providers. Thus, at every time step $t$, the solver obtains a representation for the environment, $\boldsymbol{s}_t$, and a reward $\boldsymbol{r}_t$. Based on this information, it takes an action $\boldsymbol{a}_t$ to direct the idle vehicles to different locations such that the expected discounted future reward is maximized, i.e., 
\begin{equation}
\sum_{k=t}^{\infty}\eta^{k-t}\boldsymbol{r}_{k}(\boldsymbol{a}_{t},\boldsymbol{s}_{t}), \label{total_reward_all}
\end{equation}
where $\eta<1$ is a time discounting factor. In our setting, we define the reward $\boldsymbol{r}_{k}(.)$ as a weighted sum of different performance components that reflect the objectives mentioned above. 
%
%
Since the resources (number of vehicles) are limited, some ride requests may not be served and thus we aim to minimize the difference between the demand and supply for vehicles. Also, to reduce the cost and fuel consumption, 
the number of used vehicles at any time slot $t$ should be reduced
for improved ride-sharing service. By grouping customers (and thus using more seats per vehicle), we are efficiently utilizing the fuel, reducing air pollution and alleviating the traffic congestion as less number of vehicles are used. Therefore, ride-sharing is beneficial for drives, riders, and community. \textcolor{black}{ Interaction of agent and environment is captured in Figure \ref{RL_agent}. The agent takes an action $\boldsymbol{a}_t$ based on the environment state  $\boldsymbol{s}_t$ and the corresponding reward $\boldsymbol{r}_t$.}

\section{Proposed Framework Design}

In this section, we develop DeepPool framework to solve the vehicles dispatch and routing problems. We first mathematically characterize the objective function and its components (Section~\ref{sec:objective}). Subsequently, we propose our approach, a distributed policy-- DeepPool which is a model-free approach where future demands and customer statistics  are learned to efficiently solve the dispatch and routing problems using deep Q-Network for each vehicle in a distributed manner \cite{lecun2015deep,mnih2015human}.  The reward will be learnt from the environment for individual vehicles and then leveraged by system optimizer to optimize DeepPool framework. 

\subsection{Objective}\label{sec:objective}
 In Section \ref{sec_probForm}, we define our performance metrics and briefly described each one. Here, we elaborate more and explain how every metric can be 
 used by the RL DeepPool to efficiently dispatch the vehicles. Specifically, we describe the overall centralized objective function of a dispatcher.  

The decision variables are--i) the dispatching of an available vehicle $n$ $u^{n}_{t,j}$  to  zone $j$ at time slot $t$, ii) if a vehicle $n$ is not full, decide $\zeta_{n,t}$ whether the vehicle will be available for serving new customers at the time slot $t$. $\zeta_{n,t}$ is a binary decision variable which is $1$ if it decides to serve customers, otherwise it is $0$. If the vehicle $n$ is full, naturally, $\zeta_{n,t}=0$. If vehicle $n$ is empty, it will serve the new passenger if the passenger's request generates within the region the vehicle $n$'s location at time $t$. 

%

The first component in the objective function aims to minimize the difference between the demand and supply for vehicles. Let $\boldsymbol{v}_{t,i}$ denote the number of available vehicles. Mathematically, this difference within time $t$ at zone $i$ can be written as $(\boldsymbol{v}_{t,i}-\boldsymbol{\overline{d}}_{t,i})$, and hence we need to minimize this difference for all zones, i.e., 
\begin{equation}\label{eq:mismatch}
\text{diff}^{(D)}_t=\sum_{i=1}^{M}\left(\overline{\boldsymbol{d}}_{t,i}-\boldsymbol{v}_{t,i}\right)^{+}
\end{equation}
Recall that $\overline{\boldsymbol{d}}_{t,i}$ is the estimated demand at time $t$ and zone $i$, and $\boldsymbol{v}_{t,i}$ is the number of active vehicles at time $t$ and zone $i$. Naturally, 
\begin{align}
v_{t,i}=\sum_{n}\zeta_{n,t}\mathbbm{1}\{\text{vehicle n is at zone i}\}\nonumber
\end{align} 
Note that whether the available vehicle n is at zone $i$ depends on the location of the vehicles $n$. Minimizing the demand-supply mismatch is profitable to the ride-hail services since the vehicles can serve more demand at any zone $i$. \textcolor{black}{For instance, in Figure \ref{rideSharing_Example}, assume that the vehicles have no requests at time $t_0$ and the ride requests appear at sometime in the future, say $t_5$. Hence, to minimize demand-supply mismatch, it is more likely for our policy to dispatch vehicles $1$ and $2$ to the zones where future customers are anticipated (e.g., $D, G, I$ and $H$).} 
Note that the number of available vehicles evolve randomly and inherently depends on the action variables. For example, if a vehicle drops off a passenger at a certain time (say, $\tau$) in the future at zone $j$ and it may decide to serve new users. Thus, the vehicle is available for dispatch at zone $j$ and time $\tau$. 

The demand is only picked up by the available vehicles within the zone. Thus, if the number of available vehicles within the zone $i$ exceeds the predicted demand, the vehicles can serve the demand. 

The second component of the objective function  is to minimize the 
dispatch time which refers to the expected travel time to go to zone $j$, i.e., $h^{n}_{t,j}$ if the vehicle $n$ is dispatched to zone $j$. The above time can be found from the location of the vehicle $n$ at time $t$ which is already included in the state variable $\boldsymbol{X}_{t,n}$. Note that we can only dispatch the available vehicles, i.e., if the vehicle is empty or the vehicle is partially filled and decides to serve a new user. Note that a vehicle may also go to a zone even if it does not have any new request, however, only to pick up new customers in the future by anticipating a high demand. While dispatching, one needs to minimize the dispatch time to pick up the new potential customers or go to a new zone in order to pick up a potential new customer. Thus, for all available vehicles within time $t$, we wish to minimize the total dispatch time, $T^{(D)}_{t}$,
\begin{equation}\label{eq:cruise}
T^{(D)}_{t}=\sum_{n=1}^{N}\sum_{j=1}^{M}h^{n}_{t,j}u^{n}_{t,j} 
\end{equation}
Recall that $u^{n}_{t,j}$ is the control variable which denotes the vehicle $n$ we dispatch which is available to zone $j$ at time slot $t$. 

Additionally, due to participating in ride-sharing, an extra travel time may be incurred for the users who are already in the vehicle. A user  needs to spend an additional time because of the carpooling. The time increases because of two components: i) the time required to take a detour to take new customer or a potential future customer, and ii) if  new customers are added a new optimized route is obtained based on the destinations of the customers, an additional time may be required to serve the new customers. Note that if a vehicle accepts a new user, it decides the optimal shortest time route to drop off the passengers. Hence, the time for existing passengers in the car may increase. 

Thus, we need to minimize the difference between the times a vehicle would have taken if it only serves that user and the time taken in the carpooling.   For vehicle $n$, rider $\ell$, at time $t$, we need to minimize 
$\delta_{t,n,\ell}=t^{\prime}+t_{t,n,\ell}^{(a)}-t_{n,\ell}^{(m)}$, where $t_{t,n,\ell}^{(a)}$ is the updated time the vehicle will take to drop off passenger $\ell$ because of the detour and/or new customer is added from the time $t$.  The new passenger is added $t_{t,n,\ell}^{(m)}$ is the  travel time that would have been taken if the car would have only served the user  $\ell$.   Note that if the vehicle $n$ is only serving the passenger $\ell$, then,  $t_{t,n,\ell}^{(a)}-t_{n,\ell}^{(m)}=0$.   The time $t_{t,n,\ell}^{(a)}$ is updated every time a new customer is added or an active vehicle takes a detour to travel to a new zone to take a  potential new customer. $t^{\prime}$ is the time elapsed after the user $\ell$ has requested its ride. The total extra travel overhead time, $\boldsymbol{\Delta}_{t}$, can be expressed as follows:
\begin{equation}\label{eq:delt}
\boldsymbol{\Delta}_{t}=\sum_{n=1}^{N}\sum_{\ell=1}^{U_{n}}\delta_{t,n,\ell}
\end{equation}
Note that the vehicle $n$ may not know the destination of the passenger. In that case, it will take the expected time of the passenger's travel destination generated in any region. Hence, $\delta_{t,n,\ell}$ is assumed to be mean value. 

The above time should be minimized since a user needs to spend additional time because of the carpooling which may lead to the dissatisfaction of the users. \textcolor{black}{To further clarify, let us assume that vehicle $2$, in the example depicted by Figure \ref{rideSharing_Example}, decides to serve Ride 1, Rider 2, Rider 3, and Rider 4. Thus, Rider 1 will encounter some extra trip time due to the detour to serve other Riders. Further, Rider 2 and Rider 3 will also incur additional delay if vehicle $2$ decides to drop Rider 1 and Rider 4 first. However, this overhead can be kept within a limit by tuning $\Delta_{t}$.
}

Note that if a vehicle $n$ serves only rider $\ell$, then we assume this $n^\text{th}$ vehicle uses the shortest path and thus $\delta_{t,n,\ell}=t_{\prime}$ which is the total time it takes from the request time to the servicing time. However, if carpooling is adopted, an additional travel time may be added since the vehicle has to take a detour to pick up one or more customers and the vehicle may take additional time to reach destinations of other users before dropping off a certain user.  Note that we use the shortest path routing to determine the time a vehicle would take to drop off customers. The vehicle may not drop off the passengers sequentially rather it will cover the destinations in a manner to minimize the total distance covered. Note that the optimal route can be easily obtained.
We note that in  \cite{oda2018movi}, there is no provision of car-pooling, hence a system optimizer does not need to consider the objective $\boldsymbol{\Delta}_{t}$ as it is always zero. However, we have to judiciously select the action in order to minimize the objective. The action will also depend on the distribution of the destinations of the potential customers towards the zone a vehicle is heading.


 We also want to minimize the number of used resources/vehicles. Higher number of un-used vehicles indicates that the passengers will be served using  a lower number of cars (resources). Hence, it will save the cost of  fuel cost and traffic congestion. The total number of used vehicles at time $t$ is given by
\begin{equation}
\boldsymbol{e}_{t}=\sum_{n=1}^{N}[\max\{\boldsymbol{e}_{t,n}-\boldsymbol{e}_{t-1,n},0\}] \label{activeVehicles}
\end{equation}
where $e_{t,n}$ denotes whether a vehicle is non-empty. Specifically, we want to minimize the number of vehicles which will become active from being inactive. 

Thus, the reward at time $t$ is--
\begin{align}
\overline{r}_t=-[\beta_1\text{diff}^{(D)}_{t}+\beta_2T^{(D)}_{t}+\beta_3\boldsymbol{\Delta}_{t}+\beta_4\boldsymbol{e}_{t}]
\end{align}
Note that weights $\beta_1,\beta_2, \beta_3, \beta_4$ depend on the weight factors on the objectives. Note from (\ref{total_reward_all}) that we maximize the discounted reward over a time horizon. The negative sign indicates that we want to in fact minimize the terms within the bracket. 

%
%
\begin{figure}[t]
	\includegraphics[trim=0.50in 0.85in 1.05in 1.1in, clip,width=0.45\textwidth]{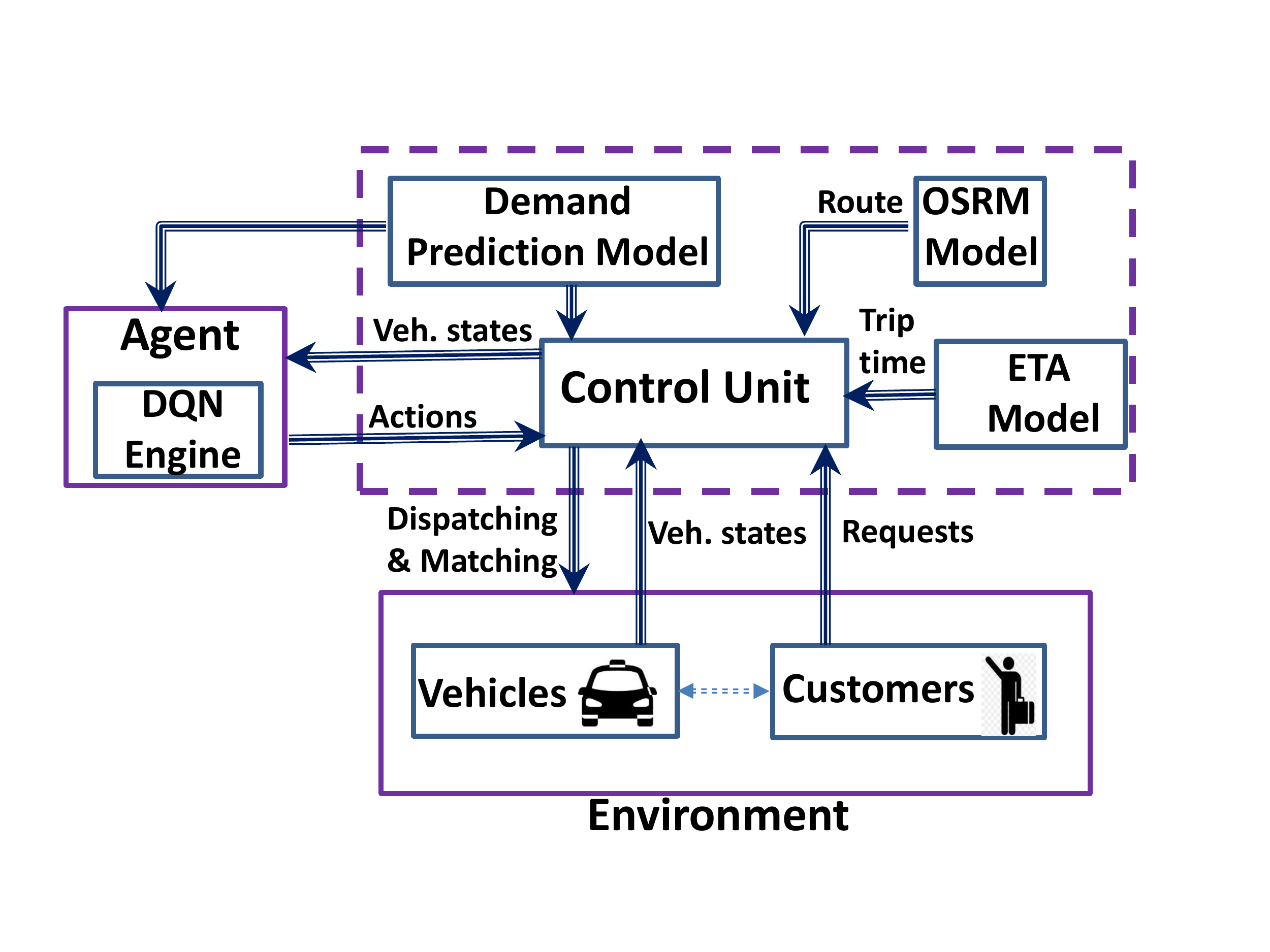}
		\vspace{-.2in}
	\caption{\textcolor{black}{DeepPool Architecture. }}
	\label{DP_arch}
	\vspace{-.2in}
\end{figure}

\subsection{DeepPool framework}

\textcolor{black}{We build a simulator to train and evaluate DeepPool framework. Figure \ref{DP_arch} shows the basic blocks of DeepPool. The different blocks and the communications between them are depicted in the figure. The control unit is responsible for maintaining the states of all vehicles, e.g., current locations, destinations, occupancy state, etc. These states are updated in every time step based on the dispatching and assignments decisions.}
Our distributed DQN policy learns the optimal dispatch actions for each vehicle. To do so, at every time slot $t$, vehicles decide $sequentially$ the next move (where to go) taking into consideration the locations of all other nearby vehicles at time slot $t$. However, their current actions do not anticipate the future actions of other vehicles. We note that it is unlikely for two (or more) drivers to take actions at the same exact time since drivers know the location updates of other vehicles in real times, through the GPS for example. {\em The advantage of the distributed DQN policy stems from the fact that each vehicle can take its own decision without coordination with other vehicles.}

\subsubsection{State}
The state variables are defined to reflect the environment status and thus affect the reward feedback of different actions. We have already defined a three tuple to capture the system updates at time $t$: $\left(\boldsymbol{X}_t, \boldsymbol{V}_{t:t+T}, \boldsymbol{D}_{t:t+T} \right)$.  


All the state elements, combined in one vector $\boldsymbol{s}_t$, can be gathered from the environment when a set of new ride requests are arrived at the DeepPool engine. For making a decision, the three-tuple state variables in $\boldsymbol{s}_t$ are pushed into the neural network input layer. 

\subsubsection{Action}
The action of vehicle $n$ is  $\boldsymbol{a}_{t,n}$. The action of a vehicle consists of two components--i) if the vehicle is partially filled it decides whether to serve the existing users or serve new user, and ii) if it decides to serve a new user or the vehicle is empty, it decides the zone the vehicle should head to at time slot $t$ which we denote by $u_{t,n,i}$. Here, $u_{t,n,i}=1$  if the vehicle decides to serve a new user and decides to go to zone $i$, otherwise it is $0$. Note that if vehicle $n$ is full it can not serve any additional user. On the other hand, if a vehicle decides to serve existing users, it uses the shortest optimal route for reaching the destinations of the users. 

\subsubsection{Reward} 
We now represent the reward  fully, and the weights which we put towards each component of the reward function for each vehicle if it is not full. First, note that  the reward is $0$ if the vehicle decides to serve the existing passengers in it (if, any). We, thus, turn our focus on the scenario where the vehicle decides to serve a new user and it is willing to take a  detour at time $t$. The reward  $r_{t,n}$ for vehicle $n$ at time slot $t$ in this case is given by
\begin{align}
r_{t,n} & =r(\boldsymbol{s}_{t,n},\boldsymbol{a}_{t,n})\nonumber\\
& =\beta_{1}b_{t,n}-\beta_{2}c_{t,n}-\beta_{3}\sum_{\ell=1}^{U_{n}}\delta_{t,n,\ell}-\nonumber\\
& \beta_{4}[\max\{e_{t,n}-e_{t-1,n},0\}\label{reward_veh_n}
\end{align}
where  $\beta_i$ is the weight for component $i$ in the reward expression. $b_{t,n}$ denotes the number of customers served by the vehicle $n$ at time $t$. $c_{t,n}$ denotes the time taken by vehicle $n$ to take a detour to pick up extra customers if the vehicle has available seats. $\delta_{t,n,\ell}$ denotes the additional time vehicle $n$ takes because of carpooling compared to the scenario if the vehicle would have served user $\ell$ directly without any carpooling. Note that while taking the decision the vehicle $n$ exactly knows the destination of the passenger. Note that when a user is added the route is updated for dropping the passengers. $U_n$ is the total number of chosen users for pooling at vehicle $n$ till time $t$. Note that $U_n$ is not known apiori which will be adapted dynamically in the DQN policy. $U_{n}$ will also vary as the passengers are picked or dropped by the vehicle $n$.  \textcolor{black}{In Figure \ref{rideSharing_Example}, $U_4$ will be equal to $4$ if vehicle $2$ decided to serve all ride requests.
}. 
The last term captures the vehicle status where $e_{t,n}$ is set to one if empty vehicle $n$ becomes occupied (even if by one passenger), however, if an already occupied vehicles takes a new customer it is $0$. The intuition behind the reward term is that if an already occupied vehicle serves a new user, the congestion cost and fuel cost will be less rather than a empty vehicle serves the user.  
Note that if we make $\beta_3$ very large, we resort to the scenario where there is no carpooling. Since high $\beta_3$ indicates that passengers will not prefer detours to serve another passenger. Thus, the setting becomes similar to the one in \cite{oda2018movi}.

We now describe how the reward function is analogous to the objective function described in Section~\ref{sec:objective}. The first term $b_{t,n}$ is related to (\ref{eq:mismatch}) since when vehicle serves more requests the mismatch between supply and demand is minimized. Also note that 
\begin{align}
b_{t,n}=\sum_{i=1}^{M}u^{n}_{t,i}\mathbbm{1}\{v_{t,i}<d_{t,i}\}\nonumber\\
\sum_{i=1}^{M} u^{n}_{t,i}=1\quad u_{t,n,i}\in\{0,1\}.
\end{align}
 \textcolor{black}{For example, in Figure \ref{rideSharing_Example},  $u_{t,2,D}=1$ if vehicle $2$ decides to go to zone $D$ to serve Rider $1$ at time $t$.
}
The second term $c_{t,n}$ in (\ref{reward_veh_n}) has an analogy with (\ref{eq:cruise}) since
\begin{align}
c_{t,n}=\sum_{i=1}^{M}h^{n}_{t,j}u^{n}_{t,j}
\end{align}
Thus, summing over $n$ returns the expression in (\ref{eq:cruise}).

The third term in (\ref{reward_veh_n}) is analogous with (\ref{eq:delt}) since it is the sum of $\sum_{\ell=1}^{U_{n}}\delta_{t,n,\ell}$ over the vehicles. Finally, the fourth term in (\ref{reward_veh_n}) is analogous with (\ref{activeVehicles}) since it determines the number of vehicles that have been active from being inactive in the previous time. {\em Though we like to make the number of active vehicles to be zero, if the total distance or the total trip time of the passengers increase, it is beneficial to use an unoccupied vehicle.}

\subsubsection{Learning Reward Function and Transition Probability}
Note that the reward stated in (\ref{reward_veh_n}) is {\em not} explicitly defined in action. The probabilistic dependence between the action and the reward function needs to be learnt. DeepPool {\em model free} approach means that the above function can be learnt using Neural network. Hence,  we learn $\mathbb{\mathbb{P}}(\boldsymbol{r}_{t}|\boldsymbol{a}_{t,}\boldsymbol{s}_{t})$
over time by feeding the current states of the system and getting the updated reward, Q-value. 
We need also to know the transition probability of the total number of active vehicles at each zone, all the vehicles' positions, and the total demand at each zone for each action taken by the vehicle. Instead of assuming any specific structure, our model free approach learns the above transition probabilities dynamically. Specifically, we need to learn
\begin{align}
\mathbb{\mathbb{P}}(\boldsymbol{s}_{t+1}|\boldsymbol{s}_t, a_{t,n}).
\end{align} 

To learn these probabilities, we use convolutional neural networks as will be  explained in Section V-B. These transition probabilities and the reward distribution determined using neural networks and used to find the Q-values in the following subsection.

\subsection{Deep Q-Network}

To make the dispatch actions in different states of the vehicles, we utilize the deep queue networks to dynamically generate optimized values. This technique of learning is widely used in modern decision making tasks due to its adaptability to dynamic features in the system. The optimal action-value function for vehicle $n$ is defined as the maximum expected achievable reward. Thus, for any policy $\pi_t$ we have 
\begin{align}
Q^{*}(s,a) & =\nonumber \\
& \underset{\pi}{\text{max}}\,\,\mathbb{E}\left[\sum_{k=t}^{\infty}\eta^{k-t}r_{k,n}\mid\left(s_{t,n}=s,\,a_{t,n}=a,\,\pi_{t}\right)\right]
\end{align}
where $0<\eta<1$ is the discount factor for the future. If $\eta$ is small (large, resp.), the dispatcher is more likely to maximize the immediate (future, resp.) reward.
At any time slot $t$,  the dispatcher monitors the current state $\boldsymbol{s}_t$ and then feeds it to the neural network to generate an action. Note that we do not use a full representation of $s_t$ in order to find the expectation. Rather, we use a neural network to approximate Q function. 

For each vehicle $n$, an action is taken such that the output of  neural network is maximized. The learning starts with zero knowledge and actions are chosen using a greedy scheme by following the Epsilon-Greedy method. Under this policy, the agent chooses the action that results in the highest Q-value with probability $1-\epsilon$, otherwise, it selects a random action. The $\epsilon$ reduces linearly from $1$ to $0.1$ over $T_n$ steps. For the $n$th vehicle, after choosing the action and according to the reward  $r_{t,n}$, the Q-value is updated with a learning factor $\alpha$ as follows
\begin{align}
Q'(s_{t,n},a_{t,n}) & \leftarrow(1-\alpha)Q(s_{t,n},a_{t,n})+\nonumber \\
& \alpha\left[r_{t,n}+\eta\,\,\underset{a}{\text{max}}\,\,Q(s_{t+1,n},a)\right]
\end{align}

Similar to $\epsilon$, the learning rate $\alpha$ is also reduced linearly from 0.1 to 0.001 over $10000$ steps. We note that an artificial neural network is needed to maintain a large system space.  When updating these values, a loss function $\mathcal{L}_{i}(\theta_{i})$ is used to compute the difference between the predicted Q-values and the target Q-values, i.e.,
\begin{equation}
\mathcal{L}_{i}(\theta_{i})=\mathbb{E}\left[\left(\left(r_{t}+\eta\,\,\underset{a}{\text{max}}Q(s,a;\overline{\theta}_{i})\right)-Q(s,a;\theta_{i})\right)^{2}\right] \label{loss_q}
\end{equation}
where $\theta_{i}$ and $\overline{\theta}_i$ are the weights of the neural networks.
This above expression represents the mean-squared error in the Bellman equation where the optimal values are approximated with a target value of $r_{t}+\eta\,\,\underset{a}{\text{max}}\,\,Q(s,a;\overline{\theta}_{i})$, using the weight $\overline{\theta}_{i}$ from some previous iterations.  

The proposed approach is shown in Algorithm \ref{alg:DeepPool}. As shown in the algorithm, we first construct the state vector $\boldsymbol{\varOmega}_{t,n}$, \textcolor{black}{using the ride requests and available vehicles, which are extracted from real records (Step 2)}. Next, for every available (not fully occupied) vehicle $n$, we select the action that maximizes its own reward, i.e., taking the {\it argmax} of the DQN-network output. The output of the DQN is the Q-values corresponding to every possible movement, while the input is the environment status governed by the vector $\boldsymbol{s}_t$ \textcolor{black}{(Steps 2 and 3)}.
Then, we update the solution by adding the tuple $(n,f_{n,t})$ to the set of dispatch solution \textcolor{black}{(Step 2)}. Finally, $\boldsymbol{X}_{t:t+T}$ is updated based on the chosen action \textcolor{black}{(line 7)}.  
We note that the updates is performed sequentially so that vehicles can take other vehicles' actions into consideration when making their own decisions. Nonetheless, each vehicle does not anticipate the future actions of other vehicles, thus limiting the coordination among them. 

\textcolor{black}{
In Algorithm \ref{DP_alg}, we fisrt start by initializing the number of vehicles and generating some ride requests in each time step based on the read record in the dataset (lines 1 to 3). Next, 
the agent determines the actions $\boldsymbol{a}_t$ using DeepPool framework and matches the vehicle to the appropriate ride request (or requests), see lines 4 to 13. Then, the vehicles traverse to the dispatched locations using the shortest path in the road network graph. The dispatched vehicles travel to dispatched locations within the time estimated by our model (Section V-A), see lines 14 to 20.	
}	

\begin{algorithm}[t]
	\caption{DeepPool Algorithm for ride-sharing using DQN\label{alg:DeepPool}}
	\begin{enumerate}
		\item \textbf{Input:} $\boldsymbol{X}_t, \boldsymbol{V}_{t:t+T}, \boldsymbol{D}_{t:t+T}$;
		\item $\quad$\textbf{Step 1}: Construct a state vector  $\boldsymbol{\varOmega}_{t,n}=\left(\boldsymbol{X}_t, \boldsymbol{V}_{t:t+T}, \boldsymbol{D}_{t:t+T} \right)$;
		
		\item $\quad$\textbf{Step 2}: Choose ${a}_{t,n}$ such that: 
		${a}_{t,n}=\underset{a}{\text{arg\,max}}Q(\boldsymbol{\varOmega}_{t,n},a;\theta)$;
		
		\item $\quad$\textbf{Step 3}: Get the destination zone $f_{t,n}$ for vehicle $n$ based on action $a_{t,n}$;

		\item $\quad$\textbf{Step 4}: Update the dispatch solution by adding $(n,f_{t,n})$
		
		\item \textbf{endfor}
		\item \textbf{output:} Dispatch Solution
	\end{enumerate}
\end{algorithm}

\begin{algorithm}[t]
\caption{\color{black} Ride-sharing Simulator}
\label{DP_alg}
\begin{algorithmic}[1]
\State Initialize vehicles states $\boldsymbol{X}_0$.
\For{$t \in T$}
\State Get all ride requests at time $t$
\For{Each ride request in time slot $t$}
    \State \textbf{Choose} a vehicle $n$ to serve the request\State according to \eqref{reward_veh_n}.
    \State \textbf{Calculate} the dispatch time using our model \State in Section \ref{RN_ETA}.
    \State \textbf{Update} the state vector $\boldsymbol{\varOmega}_{t,n}$.
    \EndFor
    \State \textbf{Send} the state vector $\boldsymbol{\varOmega}_{t}$ to the agent.
    \State \textbf{Get} the best actions (dispatch orders) $\boldsymbol{a}_t$ from \State the the agent.    
\For{ all vehicles $n \in \boldsymbol{a}_t$ }    
     \State \textbf{Find} the shortest path to every dispatch location \State  for every vehicle $n$.
     \State \textbf{Estimate} the travel time using the model in \State  Section \ref{RN_ETA}.
     \State \textbf{Update} $\delta_{t,n}$, if needed, and generate the trajectory \State  of vehicle $n$.
 \EndFor  
  \State \textbf{Update} the state vector $\boldsymbol{\varOmega}_{t+1}$.     
\EndFor
\end{algorithmic}
\end{algorithm} 


\begin{algorithm}[t]
	\caption{\color{black} Double Q-learning with experience replay}
	\label{DoubleDQN_alg}
	\begin{algorithmic}[1]
		\State \textbf{Initialize} replay memory $D$, Q-network parameter $\theta$, and target Q-network $\theta^-$.
		\For{$e:1:Episodes$}
		\State \textbf{Initialize} the simulation
		\For{$t:\Delta t:T$}
		\State \textbf{Perform} the dispatch order
		\State \textbf{Update} the state vector $\boldsymbol{\varOmega}_{t}=$ $(\boldsymbol{X}_t,  \boldsymbol{V}_{t:T},  \boldsymbol{D}_{t:T})$.
		\State \textbf{Update} the reward $\boldsymbol{r}_t$ based on actions $\boldsymbol{a}_t$.
		\For {all available vehicles $n$}
		\State \textbf{Create} $\boldsymbol{\varOmega}_{t,n}=$ $(\boldsymbol{X}_{t,n},  \boldsymbol{V}_{t:T},  \boldsymbol{D}_{t:T})$.
		\State \textbf{Store} transition
		\State $(\varOmega_{t-1,n}, a_{t-1,n},r_{t,n},\varOmega_{t,n},c_{t,n})$
		\EndFor
		\State \textbf{Sample} random transitions 
		\State $(\varOmega_i, a_i,r_i,\varOmega_{i+1},c_{i+1})$ from $D$.
		\State Set $a_{i}^{*}=\text{argmax}_{a^{'}}Q(\varOmega_{i+1},a_{i}^{*};\theta^{-})$.
		\State Set $z_{i}=r_{i}+\gamma^{1+c_{i+1}}\hat{Q}(\varOmega_{i+1},a_{i}^{*};\theta^{-})$.
		\State \textbf{minimize} $(z_{i}-Q(\varOmega_{i},a_{i};\theta))$ w.r.t. $\theta$.
		\State Set $\theta=\theta^-$ every $N$ steps.
		\State \textbf{Update} the set of available vehicles $A_{t}$
		\For {$n$ in $A_t$}
		\State \textbf{Create} $\boldsymbol{\varOmega}_{t,n}=$ $(\boldsymbol{X}_{t,n},  \boldsymbol{V}_{t:T},  \boldsymbol{D}_{t:T})$.
		\State \textbf{Choose}, with prob. $\epsilon$,  a random action from 
		\State $a_t^{(n)}$.
		\State \textbf{Else} set $a_{t}^{(n)}=\text{argmax}_{a}Q(\varOmega_{t}^{(n)},a;\theta)$.
		\State \textbf{Send} vehicle $n$ to its destination, based on 
		\State $a_{t}^{(n)}$.
		\State \textbf{Update}  $\boldsymbol{\varOmega}_{t,n}$.
		\EndFor
		\EndFor
		\EndFor  
	\end{algorithmic}
\label{doubleQlearning}
\end{algorithm}

\color{black}


\section{DeepPool Simulator Design}

To validate and evaluate our DeepPool framework, we create a simulator based on real public dataset of taxi trips in Manhattan, New York city \cite{manh_dataSet,zhan2013urban}. We consider the data of June 2017 for training neural networks (as explained in Sections V.A, V.B), and one week from July 2017 for evaluations. For each trip, we obtain the pick-up time, location, passenger count, and drop-off locations. We use this trip information to construct travel demand prediction model. Further, we use Python and tensorflow to implement the DeepPool framework. 

\subsection{Road Network and Estimated Time Models} \label{RN_ETA}
We construct a region graph relying on the New York city map, obtained from OpenStreetMap \cite{OpenStreetMap}. Also, we construct a directed graph as the road network by partitioning the city into small service area $212\times219$ bin locations of size $150m\times 150m\,$. We find the closest edge-nodes to the source and destination and then search for the shortest path between them. To estimate the minimal travel time for every dispatch, we need to find the travel time between every two dispatch nodes/location \footnote{\textcolor{black}{The location refers to a particular place inside a zone. Different locations inside a zone could have different dispatch/trip times to a particular location in another zone, depending on latitude and longitude of the pickup/dropoff location of the rider. }}. In order to learn travel time for every dispatch, we  build a fully connected neural network with historical trip data as an input and the travel time as output. The fully connected multi-layer perception network consists of
two hidden layers with width of 64 units and rectifier nonlinearity. The output of this neural network gives the expected time between zones. To avoid over-fitting issues and to get better training, we shuffle the two weeks data (last two weeks from the June data) and then split it into two categories: 70\% for training and the remaining 30\% for testing. We found that the error for both training and testing datasets is pretty small and very close to each other, which is approximately 3.75 minutes. \textcolor{black}{While this model is relatively simple (contains only two hidden layers), our goal is to achieve a reasonable accurate estimation of dispatch times, with short running time.}
Finally, if there is no vehicle in the range of $5km^2$, the request is considered rejected. 
\textcolor{black}{
\subsection{Demand Prediction} \label{DemandPred}
We use a convolutional neural network to predict future demand. The output of the network is a $212\times219$ image such that each pixel represents the expected number of ride requests in a given zone for $30$ minutes ahead. The network input consists two planes of actual demand of the last two steps. The size of each plane is $212\times219$. The first hidden layer convolves $16$ filters of size $5\times5$ with a rectifier nonlinearity, while the second layer convolves $32$ filters of $3\times3$ with a rectifier nonlinearity. The output layer convolves $1$ filter of size $1\times1$ followed by a rectifier nonlinear function. We split the (two weeks) data such that the first $70\%$ is used for training and the last $30\%$ is used for validation. The root-mean-squared errors for training and validation are $1.021$ and $1.13$, respectively.
}

\subsection{DQN of DeepPool}

We used conventional neural networks to estimate the Q-values for each individual action. The service area is divided into $43\times44$, cells each of size  $800m\times800m$. The vehicle can move (vertically or horizontally) at most $7$ cells, and hence the action space is limited to these cells. A vehicle can move to any of the $14$ vertical ($7$ up and $7$ down) and $14$ horizontal ($7$ left and $7$ right). This results in a $15\times15$ map for each vehicle as a vehicle can move to any of the $14$ cells or it can remain in its own cell. The input to the neural network consists of the state representation, demand and supply, while the output is the Q-values for each possible action/move (15 moves). \textcolor{black}{The input consists of a stack of four feature planes of demand and supply heat map images each of size 51x51. In particular, first plane includes the predicted number of ride requests next $30$ minutes in each region, while the three other planes provide the expected number of available vehicles in each region in $0, 15$ and $30$ minutes. Before passing demand and supply images into the network, different sizes of average pooling with stride $(1, 1)$ to the heat maps are applied, resulting in $23\times23\times4$ feature maps. The first hidden layer convolves 16 filters of size 5x5 followed by a rectifier nonlinearity activation. The second and third hidden layers convolve 32 and 64 filters of size 3x3 applied a rectifier nonlinearity. Then, the output of the third layer is passed to another convolutional layer of size $15\times15\times128$. The output layer is of size $15\times15\times1$ and convolves one filter of size 1x1.
}
\textcolor{black}{
	\begin{figure}[t]
		\includegraphics[trim=2.5in 1.65in 2.50in 1.2in, clip,width=0.45\textwidth]{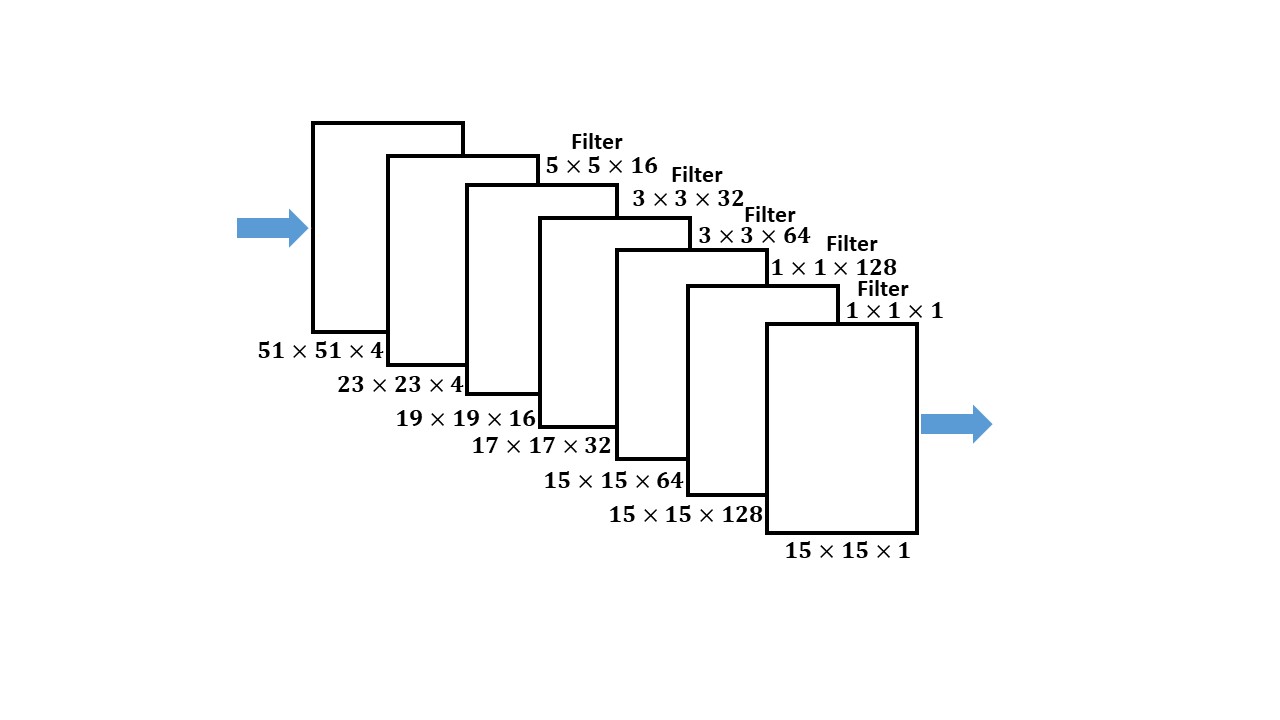}
		\vspace{-.1in}
			\caption{\textcolor{black}{ The architecture of the Q-network. The output represents the Q-value for each possible movement/dispatch.}}
		\label{Q_netArch}
		\vspace{-.3in}
	\end{figure}
}

Since reinforcement learning is unstable for nonlinear approximations such as the neural network, due to correlations between the action-value, we use experience replay to overcome this issue \cite{schaul2015prioritized}. \textcolor{black}{
Since every vehicle runs its own DQN policy, the environment during training changes over time from the perspective of individual vehicles. To mitigate such effects, a new parameter $\alpha$ is imposed to give the probability of performing an action in each time step. Here, $\alpha$ increases linearly from $0.3$ to $1.0$ over first $5000$ steps. Therefore, only $30\%$ of the vehicles takes an action at the beginning. Thus, once the DQN approaches the optimal policy, the number of vehicles moving in each time step does not vary significantly. Algorithm \ref{doubleQlearning} shows a detailed steps for the training phase.
}

\begin{figure}[ht]
	\begin{center}
		\begin{minipage}[b]{0.49\linewidth}
			\includegraphics[trim=0.1in 0.02in 4.2in .0in, clip,width=\textwidth]{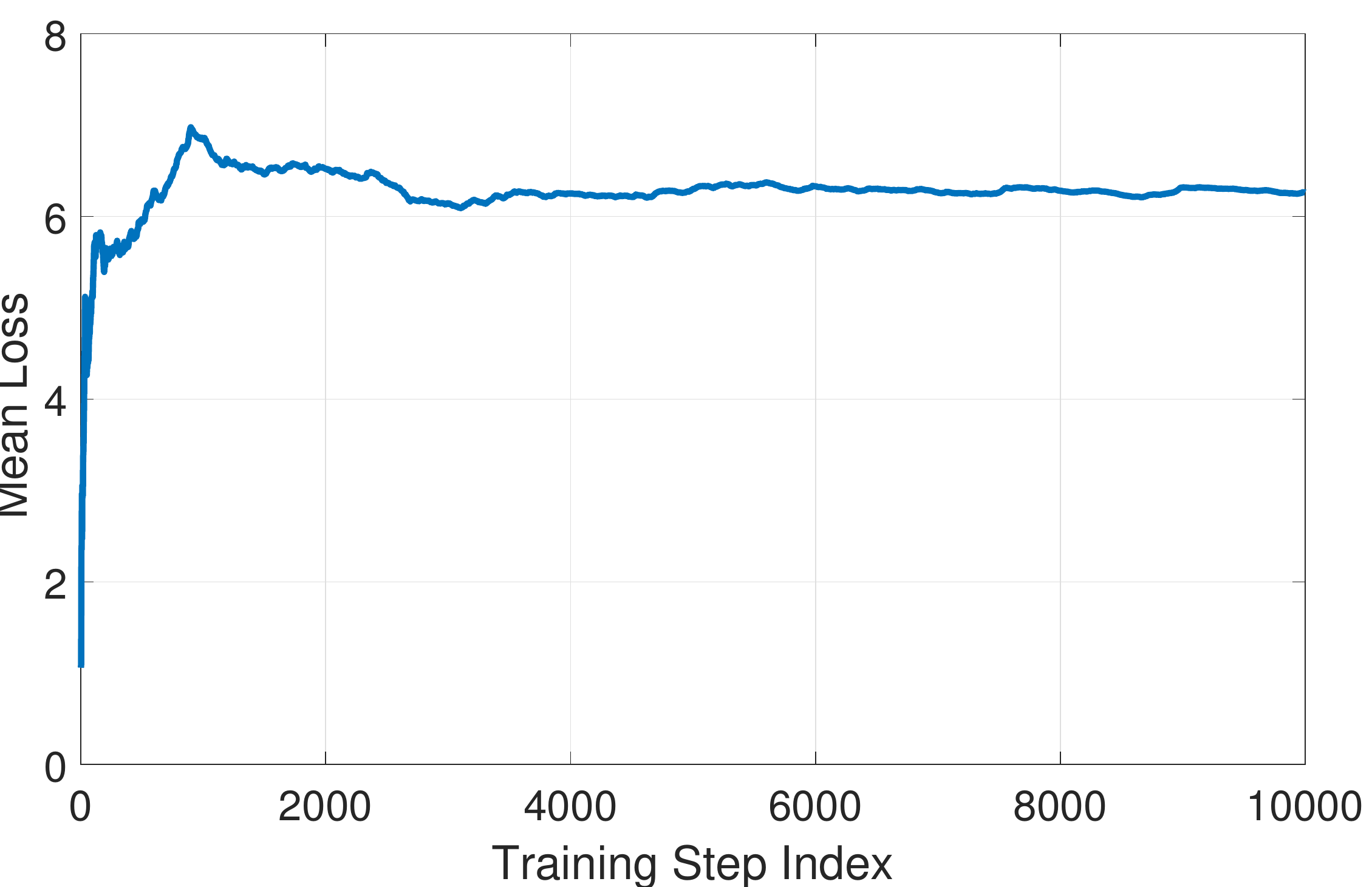}
			\vspace{-.3in}
		\caption{Average loss versus the training step index.}
		\label{avg_loss}
		\end{minipage}
	\begin{minipage}[b]{0.49\linewidth}			\includegraphics[trim=0.15in 0.061in 3.6in .0in, clip,width=\textwidth]{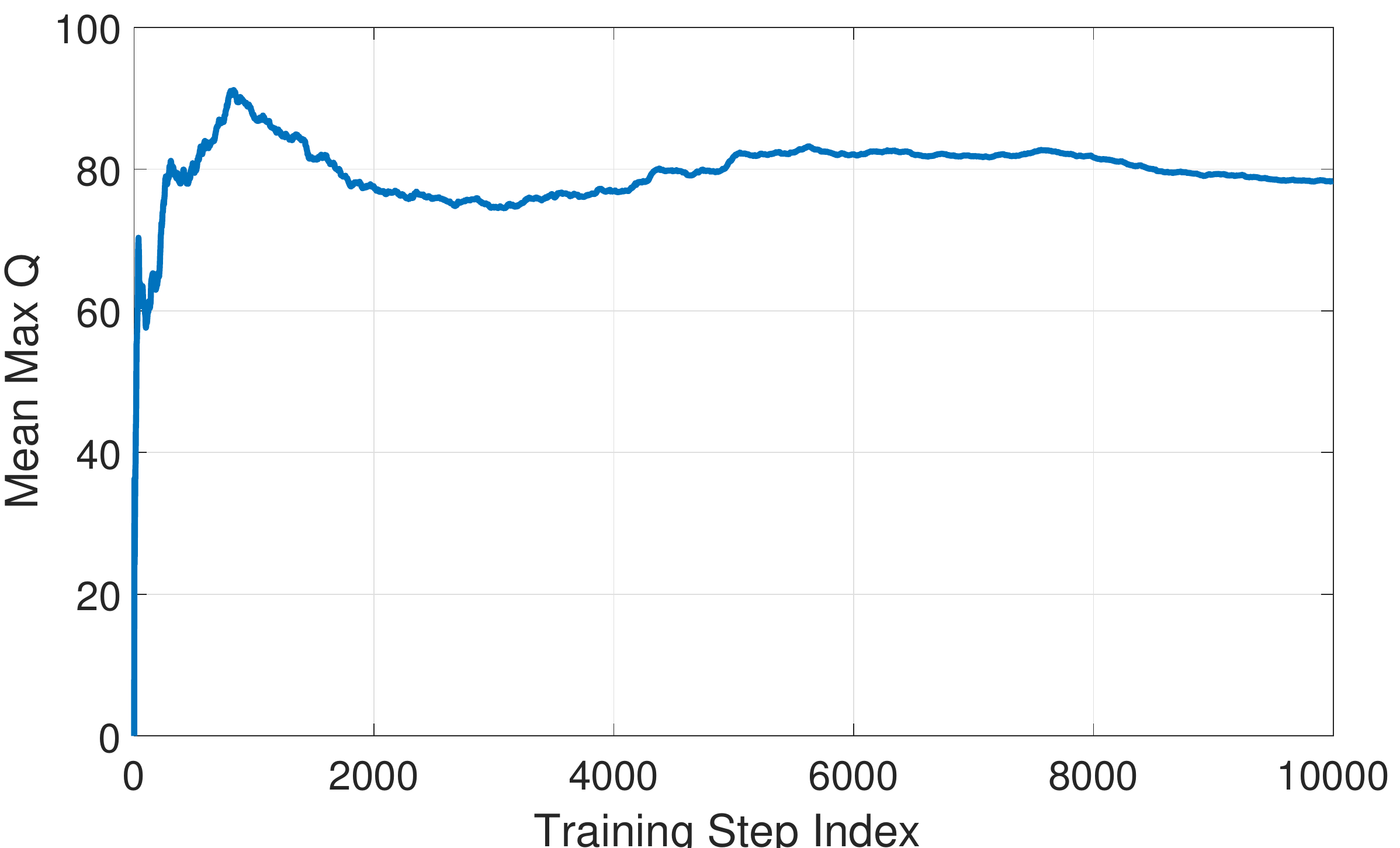}
			\vspace{-.3in}
		\caption{Average predicted action-value versus the training step index for the DeepPool agent.}
		\label{avg_max_Q}
			\end{minipage}
	\end{center}
	\vspace{-.2in}
\end{figure}

We trained our neural networks, using the data from June 2017, for 10000 epochs and used the most recent $5000$  experiences as a replay memory.  In Figure \ref{avg_loss}, we show the evolution of the average loss $\mathcal{L}_{i}(\theta_{i})$ (defined in equation \eqref{loss_q}) with the training steps. We see that the average loss variations reduces as the training steps increase, hence, the predicted Q-values converge to the target ones. Further, Figure \ref{avg_max_Q} gives the average of the maximum Q-values. We note that once the max Q-value is around 90, it starts to decrease and thus gives the vehicles more opportunities to compete on gaining customers. This decreasing average in the mean of the Q-values is because of the environment variations, which reduces the return/reward individual  vehicle can get.

\section{Evaluation Results}
In this section, we present our evaluation results based on the simulator setup described in the last section. 
We carry out numerical evaluations based on real-world dataset traces for one  week (the first week of July 2017) to evaluate the performance of our proposed DeepPool framework.

\subsection{Baselines}
We compare the proposed  DeepPool strategy with three baseline policies, as described below.
\begin{itemize}
	\item No Dispatch-No Ride-sharing (NoDS-NoRS): In this policy, we did not dispatch the vehicles and the ride-sharing is not allowed. The initial locations of the vehicles are chosen from the pickup locations of the ride requests in the dataset.

	\item No Dispatch with Ride-Sharing (NoDS-RS): In this policy, we did not dispatch the vehicles. However, ride-sharing is allowed. Similar to NoDS-NoRS, the initial locations of the vehicles are chosen from the pickup locations of the ride requests in the dataset. 
	
	\item Dispatch and No Ride-Sharing (DS-NoRS) \textcolor{black}{: This policy is akin to \cite{oda2018movi}}. In this policy, ride-sharing is not considered, but the vehicles are still dispatched. Thus, each vehicle can serve only one ride request and it cannot serve other passengers unless it drops off the current rider. Note that this policy follows as a special case of our proposed policy by allowing every vehicle to serve only one ride request. 
	\textcolor{black}{
	\item Dispatch with minimum distance with ridesharing (DS-mRS)\cite{bei2018algorithms}: In this policy, ride-sharing is allowed where two riders/customers are assigned to one vehicle so that the total driving distance is as small as possible \cite{bei2018algorithms}. Note that vehicles are still  dispatched as per our approach.}
    \textcolor{black}{
    \item Dispatch-Ridesharing with Actual Demand (DS-aRS): This policy is similar to DeepPool except that the demand is not anticipated but, rather, it is the actual demand taken from the real records in the dataset of New York City \cite{manh_dataSet}. 
    \item Centralized Receding Horizon Control (cRHC): In this policy, the dispatch actions are taken to maximize the expected reward, defined in \ref{total_reward_all} in a centralized manner. For detailed treatment of this policy, interested reader is referred to \cite{miao2016taxi,oda2018movi}
    }

\end{itemize}
\begin{figure}[b]
	\begin{center}
		\includegraphics[trim=0.5in 0.02in 1.2in .3in, clip,width=0.5\textwidth]{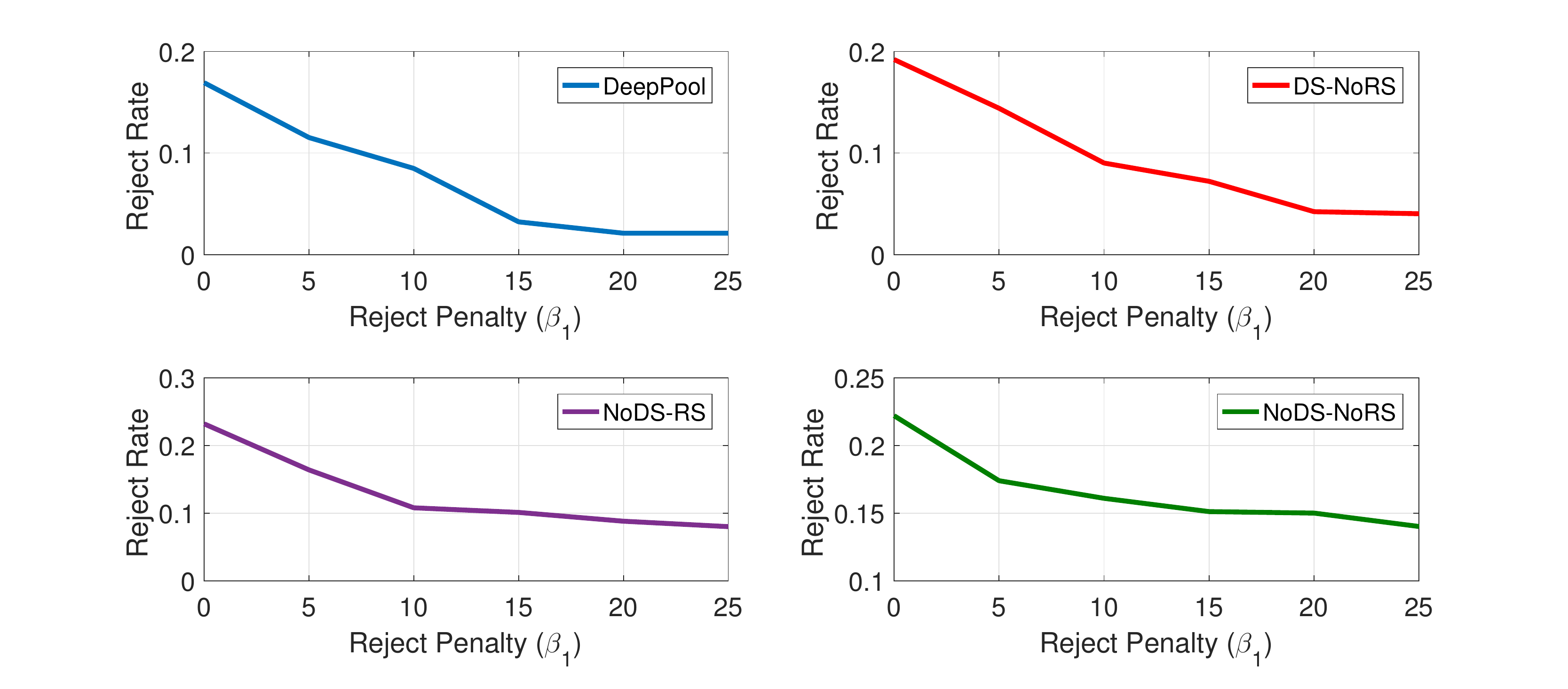}
			\vspace{-.4in}
		\caption{Average reject rate per request for different reject penalty.}
		\vspace{-.1in}
		\label{rr_vs_beta1}
	\end{center}
\end{figure}

\begin{figure*}[htbp]
		\begin{minipage}[b]{0.32\linewidth}
			\includegraphics[trim=0.25in 0.02in 4.2in .0in, clip,width=\textwidth]{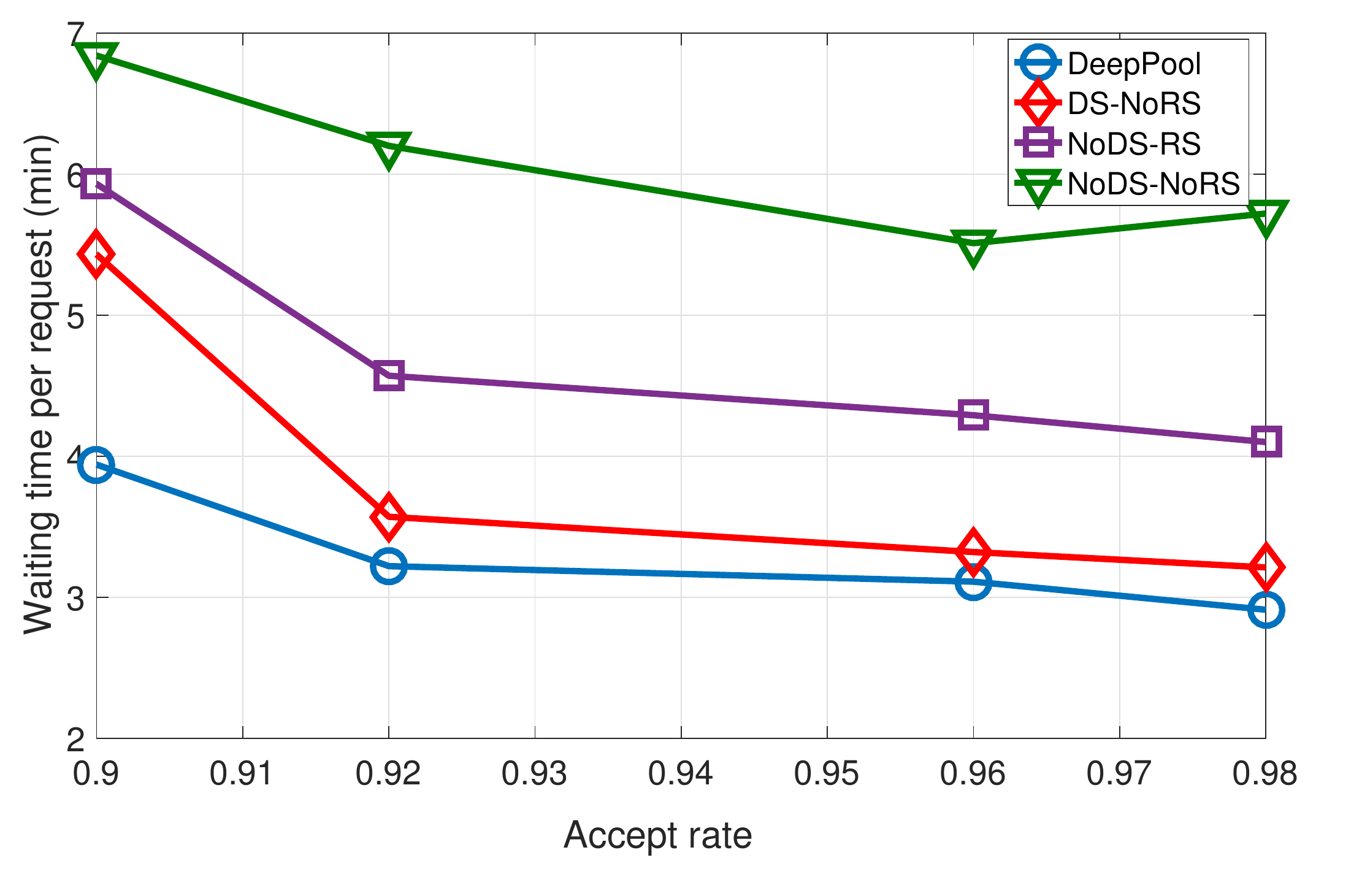}
			\vspace{-.3in}
			\caption{Average waiting time per request for different accept rate.}
			\label{waitTime_vs_beta1}
		\end{minipage}	
	\hspace{.01in}
		\begin{minipage}[b]{0.32\linewidth}
			\includegraphics[trim=0.15in 0.02in 4.2in .0in, clip,width=\textwidth]{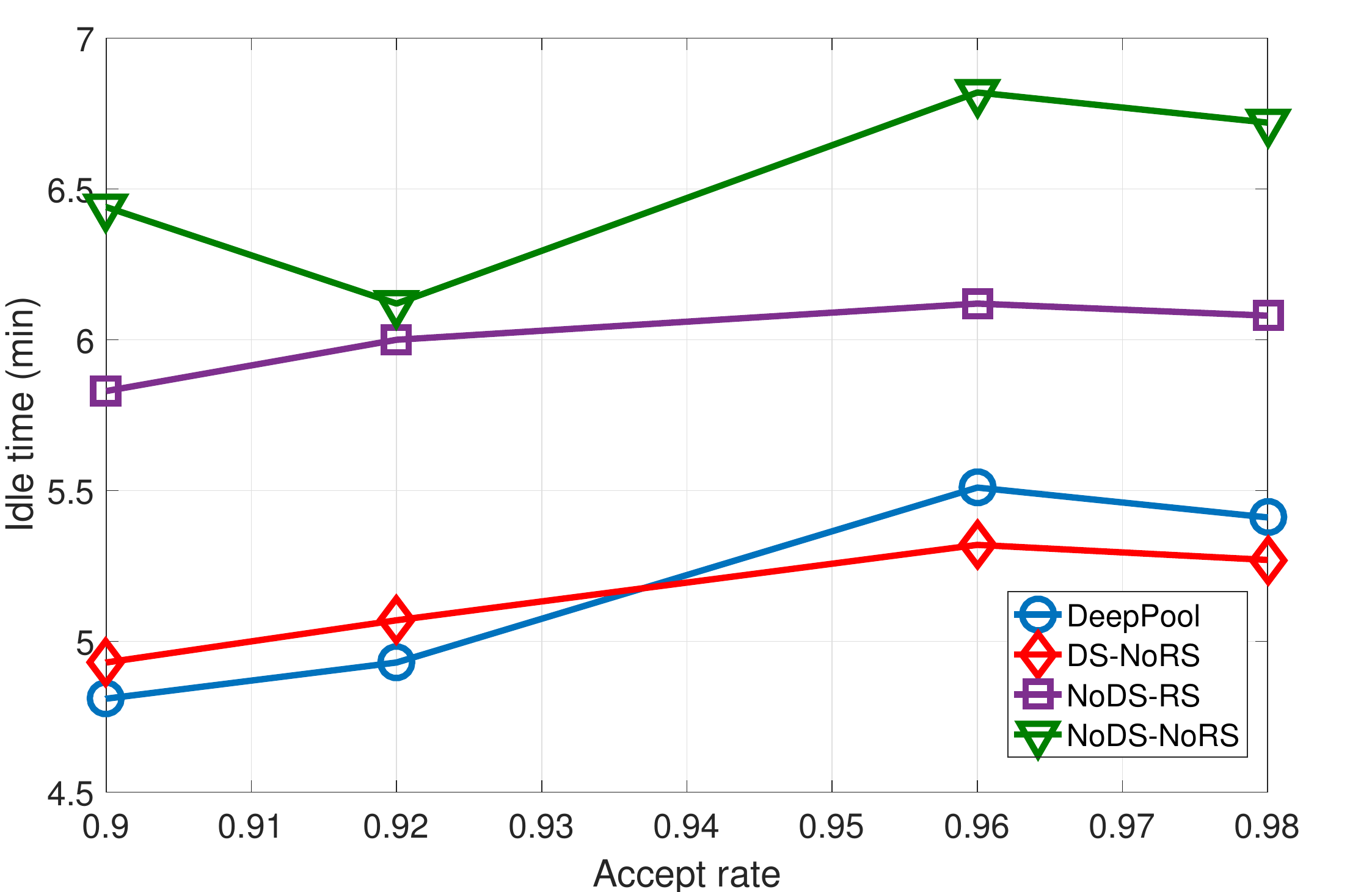}
			\vspace{-.3in}
			\caption{Average idle time per request for different accept rate.}
			\label{idleTime_vs_beta1}
		\end{minipage}
	\hspace{.01in}
		\begin{minipage}[b]{0.32\linewidth}
			\includegraphics[trim=0.25in 0.02in 4.0in .0in, clip,width=\textwidth]{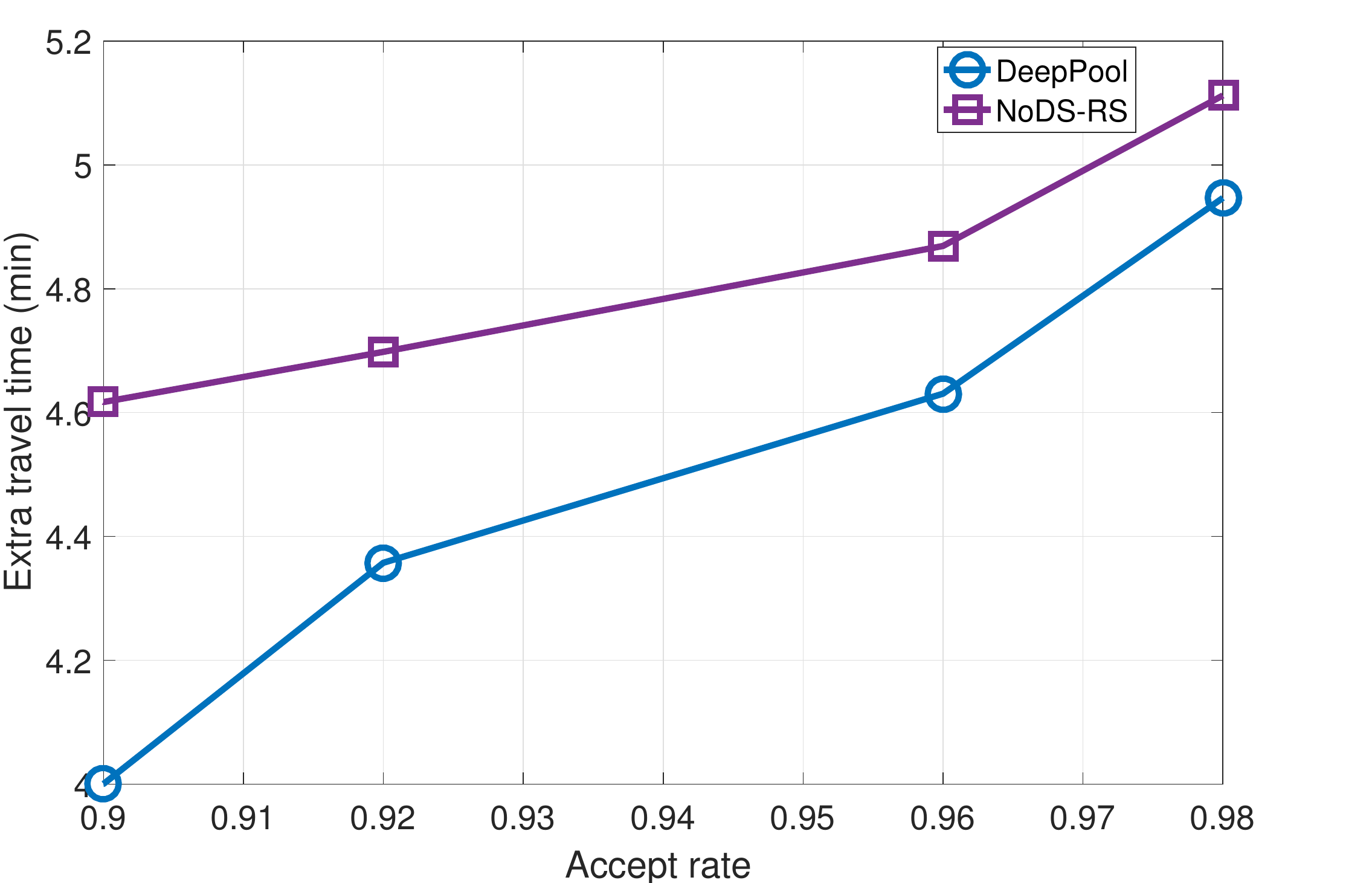}
			\vspace{-.3in}
			\caption{Average travel time overhead per request for different accept rate.}
			\label{overHead_vs_beta1}
		\end{minipage}
		\hspace{.01in}
		\begin{minipage}[b]{0.32\linewidth}
			\includegraphics[trim=0.25in 0.02in 4.0in .0in, clip,width=\textwidth]{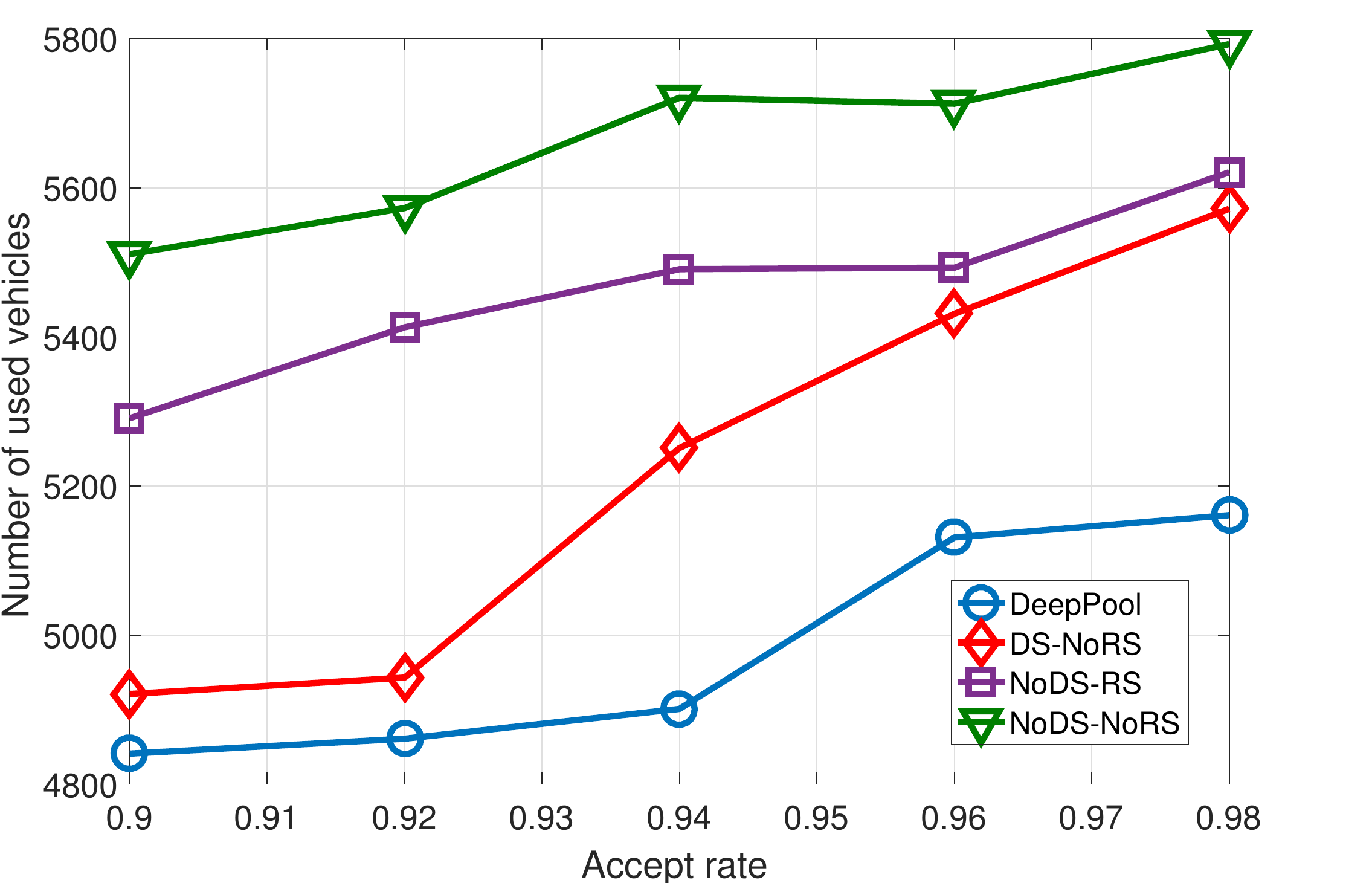}
			\vspace{-.3in}
			\caption{Average number of used vehicles for different accept rate.}
			\label{usedCars_vs_AR}
		\end{minipage}
		\hspace{.01in}
		\begin{minipage}[b]{0.32\linewidth}
		\includegraphics[trim=0.25in 0.02in 4.0in .0in, clip,width=\textwidth]{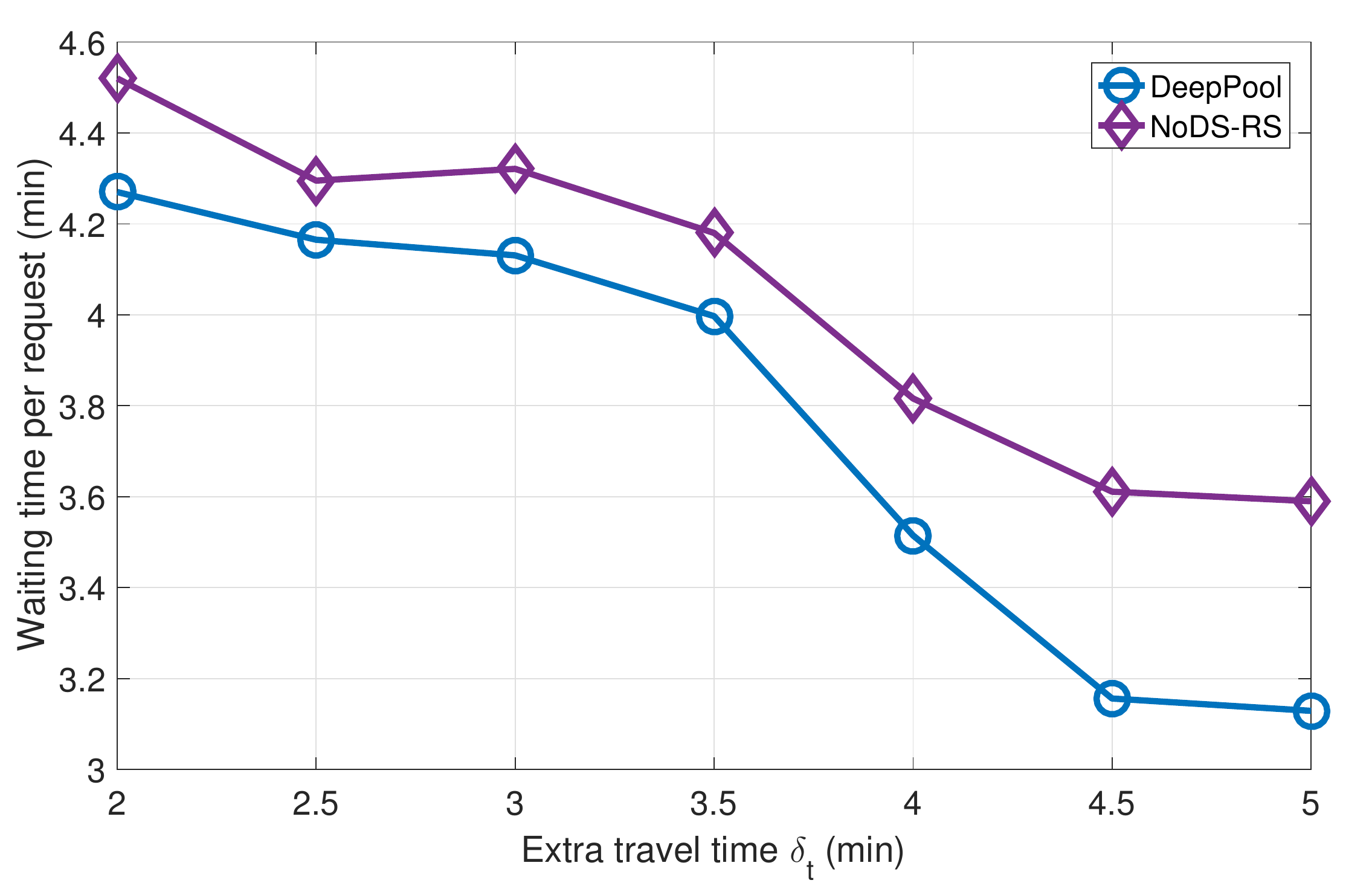}
		\vspace{-.3in}
		\caption{Average waiting time per request for different overhead values $\delta_t$.}
		\label{waitTime_vs_overhead}
	\end{minipage}	
	\hspace{.01in}
	\begin{minipage}[b]{0.32\linewidth}
		\includegraphics[trim=0.25in 0.02in 4.0in .0in, clip,width=\textwidth]{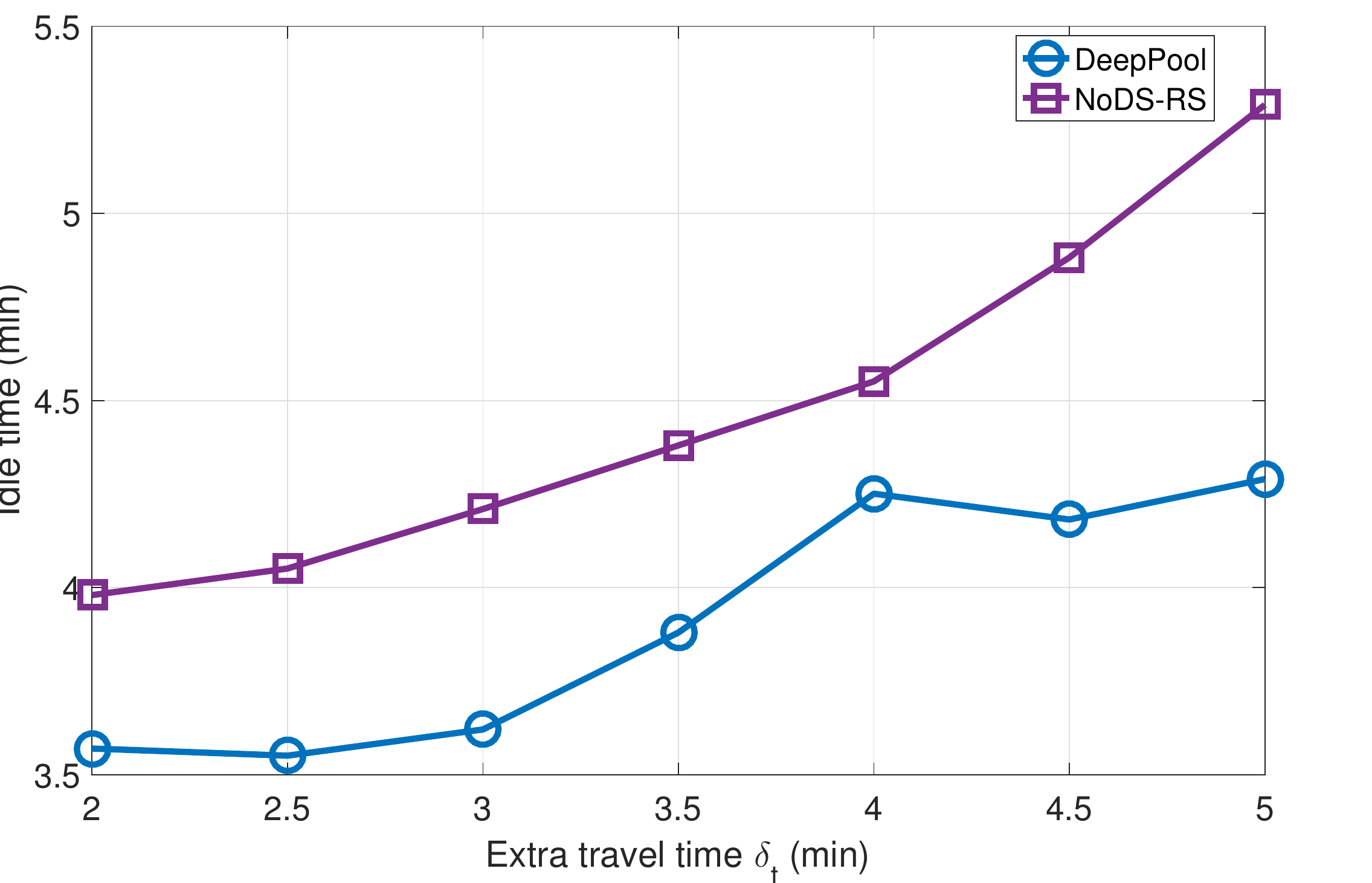}
		\vspace{-.3in}
		\caption{Average idle time per vehicle for different overhead values $\delta_t$.}
		\label{idleTime_vs_overhead}
	\end{minipage}
	\vspace{-.2in}
\end{figure*}

\subsection{Results}
To initialize the environment, we run the simulation for $20$ minutes without dispatching the vehicles. Further, we set the number of vehicles to $N=6000$. The initial locations of these vehicles correspond to the  
first $6000$ ride requests. We set the maximum horizon to $T=30$ step and $\Delta t =1$ minute. We breakdown the reward and investigate the performance of the different policies.
Further, unless otherwise stated, we set $\beta_1=10$, $\beta_{2}=1$, $\beta_{3}=5$, and $\beta_{4}=8$. 

Recall that the reward function is composed of four different metrics, which we wish to minimize: supply-demand mismatch, dispatch time, extra travel time due to ride-sharing, and number of used vehicles (resources). We note that the supply-demand mismatch is reflected in our simulation through a reject rate metric. This metric is computed as the number of ride rejects divided by the total number of requests per day. Recall that a request is rejected if there is no vehicle around in a range of $5km^2$, or no vehicle is available to serve the request. Also, the metric of idle time represents the time at which a vehicle is not occupied while still incurring gasoline cost and not gaining revenue. In addition,  the waiting time represents the time from the ride request till the time at which the vehicle arrives to pick up the customer.

\begin{figure}[t]
	\includegraphics[trim=0.15in 0.02in 4.0in .0in, clip,width=0.40\textwidth]{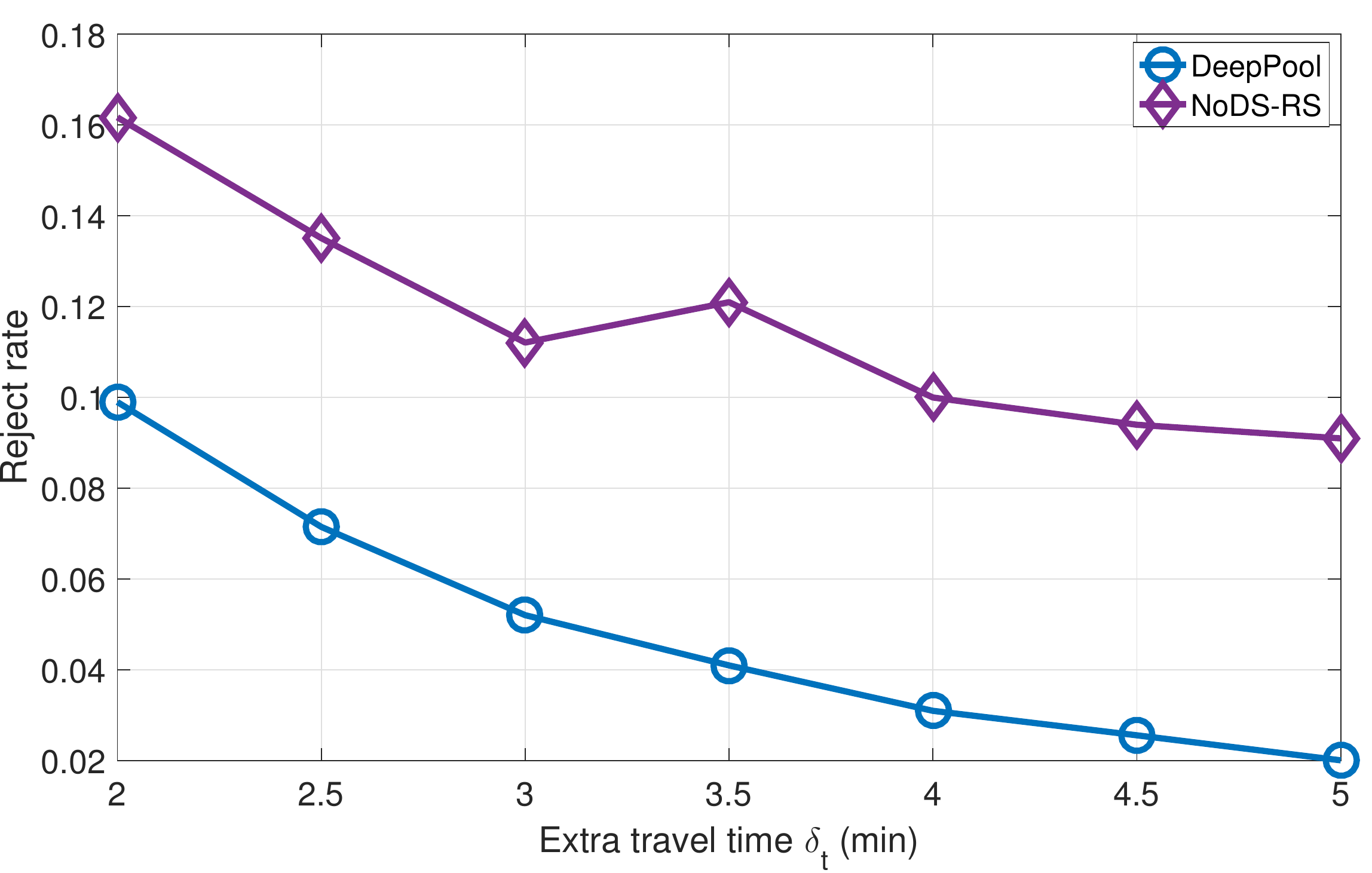}
		\vspace{-.15in}
	\caption{Average reject rate for different overhead values $\delta_t$.}
	\label{rr_vs_overhead}
		\vspace{-.3in}
\end{figure}

%
%


\begin{figure*}
		\centering
\begin{minipage}[t]{0.22\textwidth}
\includegraphics[trim=0.1in 0.02in 4.05in 0.02in, clip,width=\textwidth]{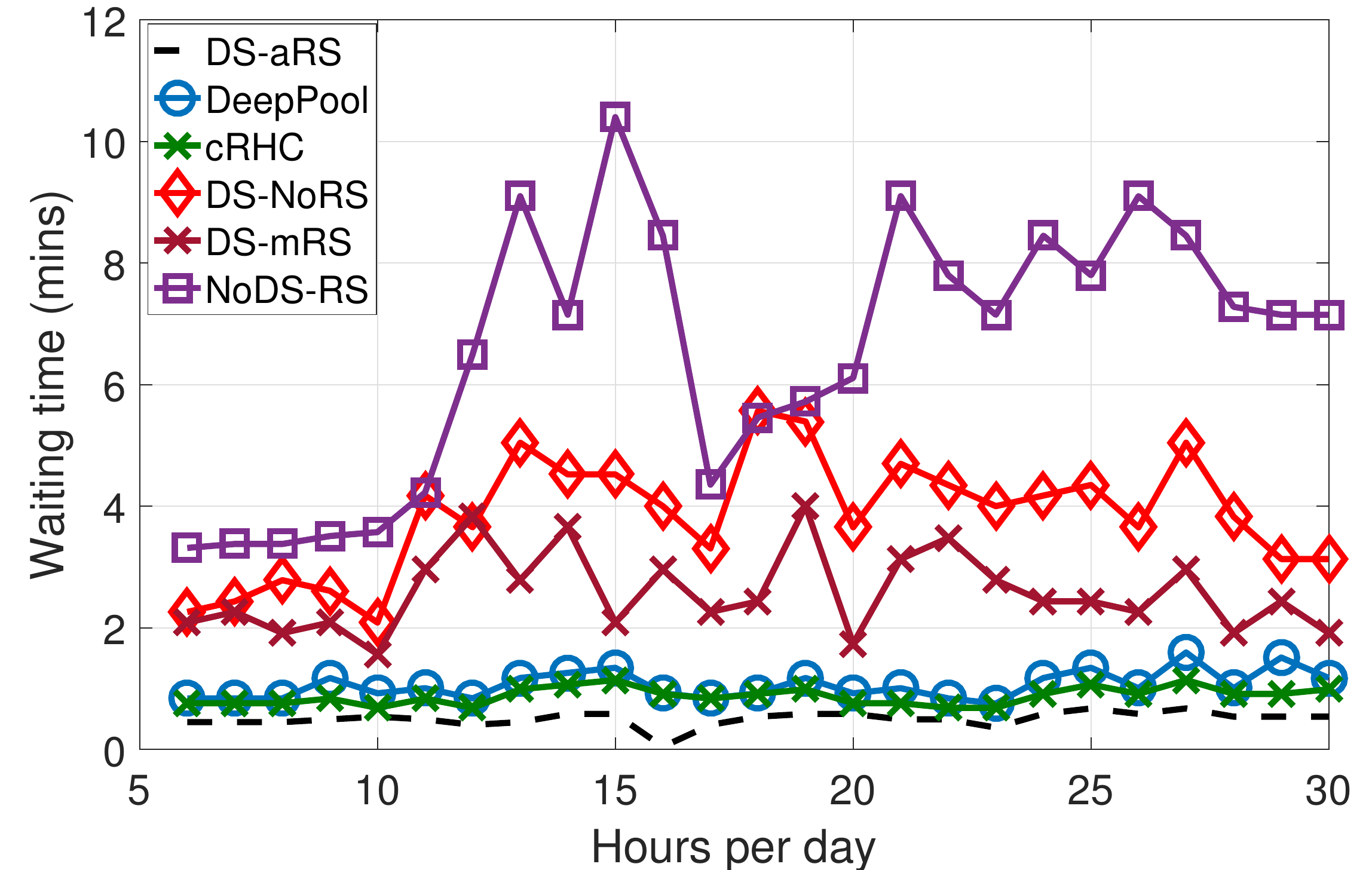}
\vspace{-.3in}
\caption{\small Waiting time comparison for the different policies on Sunday.  
	\label{waitTime_vs_waiting}}
\end{minipage}
\hspace{.01in}
\begin{minipage}[t]{0.22\textwidth}
	\centering\includegraphics[trim=0.1in 0.05in 4.05in 0.0in, clip,width=\textwidth]{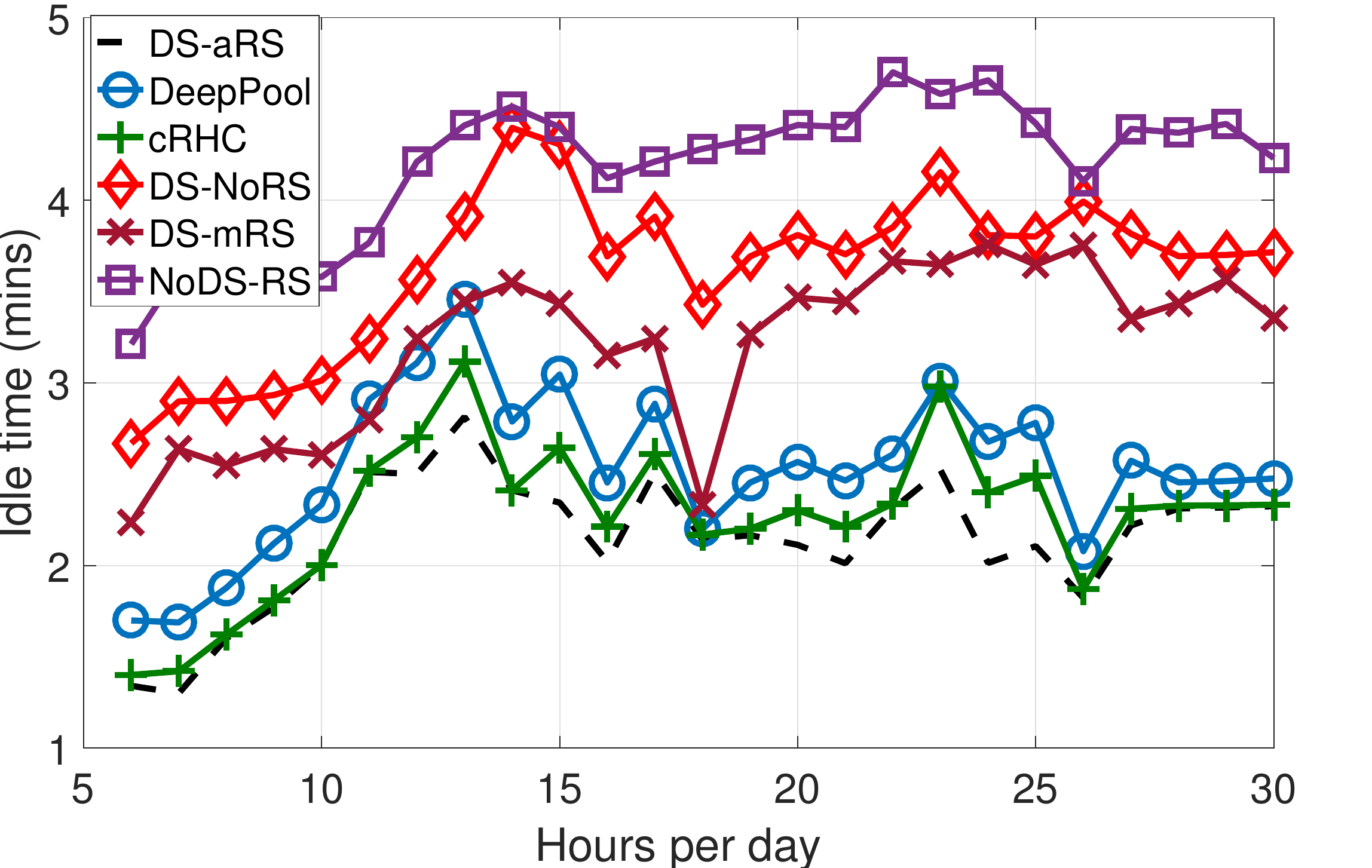}
\caption{Idle time comparison for the different policies on  Sunday.
		\label{idleTime_vs_days}}
\end{minipage}
\hspace{.01in}
\begin{minipage}[t]{0.22\textwidth}
		\centering\includegraphics[trim=0.1in 0in 4.15in 0.0in, clip,width=\textwidth]{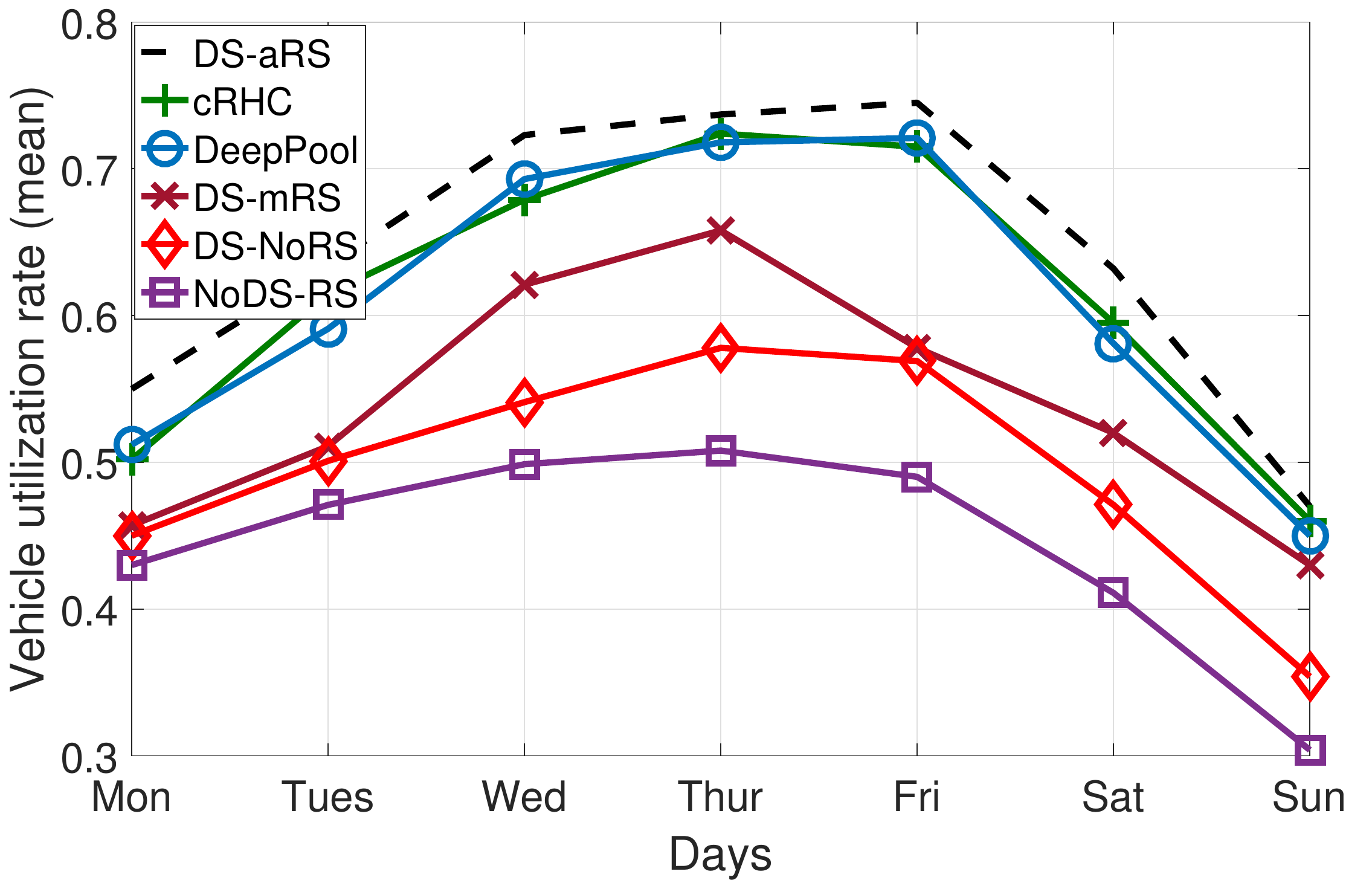}
\vspace{-.3in}
	\caption{\small Average utilization rate versus the days of the week.
		\label{Utili_mean}}
\end{minipage}
\hspace{.01in}
\begin{minipage}[t]{0.22\textwidth}
	\centering\includegraphics[trim=0.1in 0in 4.15in 0.0in, clip,width=\textwidth]{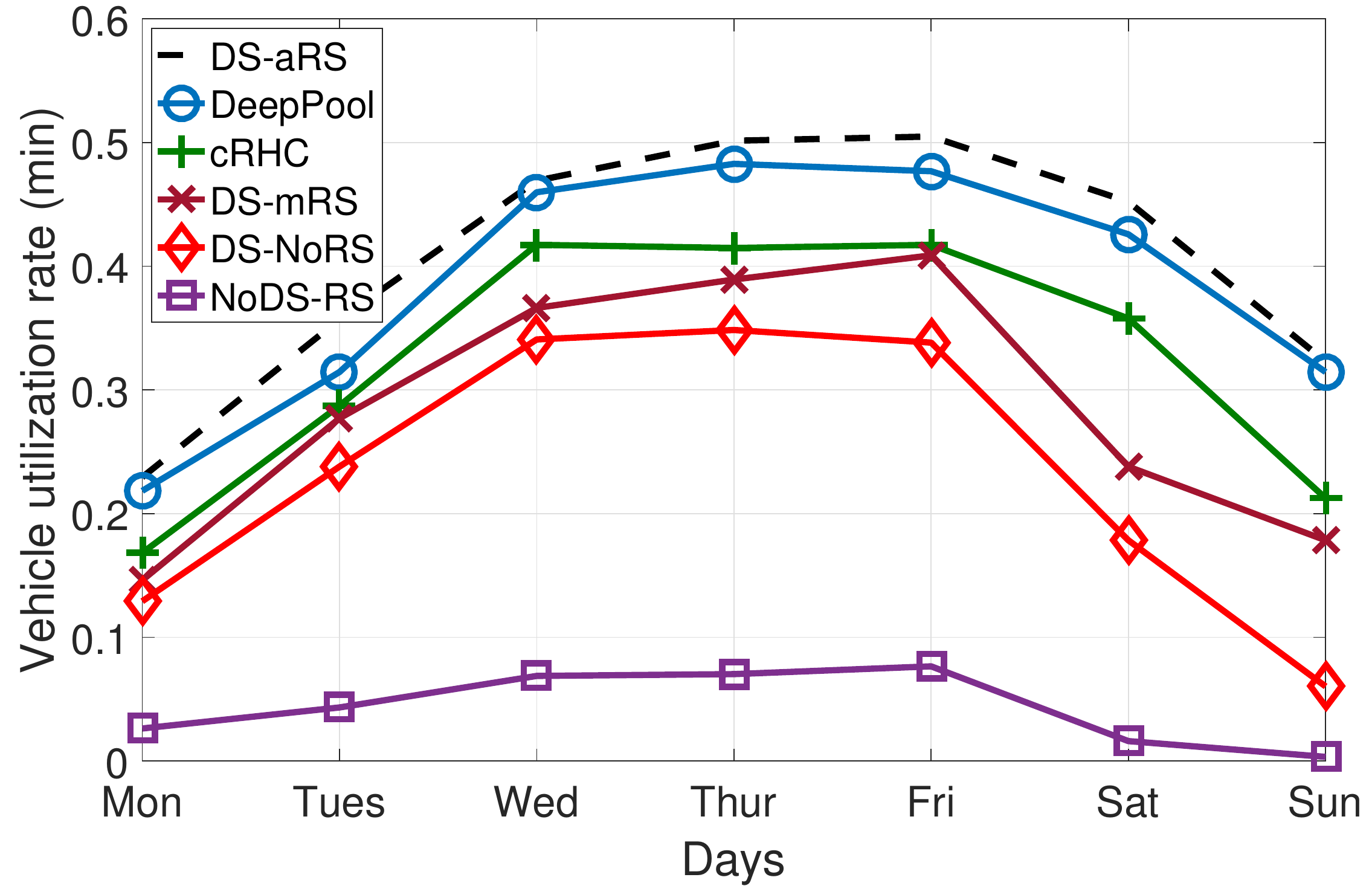}
	\vspace{-.3in}
	\caption{\small Minimum utilization rate versus the days of the week.
		\label{Utili_min}}
\end{minipage}
\vspace{-.2in}
\end{figure*}

Figure \ref{rr_vs_beta1} shows that the  reject rate is decreasing for increased $\beta_1$ for all the algorithms as a higher weight is given to minimize this metric in the reward expression.  This change of reject rate with varying $\beta_1$ is captured in Figures \ref{waitTime_vs_beta1}-\ref{usedCars_vs_AR} to evaluate the performance of different metrics with changing arrival rate. Thus, the points in Figures \ref{waitTime_vs_beta1}-\ref{usedCars_vs_AR}  are for different values of $\beta_1$ while keeping the rest of the $\beta$ values as fixed. We capture the waiting time of passengers for different accept rate (defined as 100\% - reject rate) in Figure \ref{waitTime_vs_beta1}. We can notice that both optimized dispatch and pooling can help reduce the waiting time.
This also tells that vehicles might be close to passengers locations but many requests are not served quickly due to either a vehicle cannot accommodate more than one ride (as in NoRS based policies) or not realizing the proximity of ride requests (as in the NoDS-based policies). This observation emphasizes on the need for distributed optimized ride-sharing based algorithms. 

Figure \ref{idleTime_vs_beta1} plots the idle time for different accept rate. We note that reducing the idle time is important to save the gasoline cost and to increase the utilization of  vehicles. DeepPool achieves the lowest idle time (for low accept rate regime) by intelligently dispatching the vehicles to places where future demand is more likely to occur and more than one customer can be grouped and thus being served together by the same vehicle. Thus, vehicles can find passengers to serve at reduced idle time and lower incurred cost for gasoline.

Figure \ref{overHead_vs_beta1} plots the extra travel time for different accept rate. Since DS-NoRS and NoDS-NoRS policies do not have ride-sharing flexibility, and thus having zero extra time, $\delta_t=0$, we do not plot them to highlight the differences between the two ride-sharing based approaches (NoDS-RS and DeepPool). The extra time increases when accept rate increases since more customers are being pooled together, thus vehicles behave in a greedy manner to get more customers. However, this would increase the idle time of some other non-used vehicles as depicted in Figure \ref{idleTime_vs_beta1}.

Figure \ref{usedCars_vs_AR} shows the number of used vehicles with respect to the accept rate. Clearly, DeepPool utilizes the vehicles capacity better and thus reduce the number of active vehicles. Thus, our approach is able to serve the demand with less number of cars as compared to other approaches, thanks to pooling capability and optimized dispatching policy. We also observe that optimizing only the  dispatching of vehicles does not reduce much the number of used cars especially in the regime where the high acceptance rate is high. Hence, our policy can reduce the traffic and fuel consumption as less resources in needed per unit time.

Seeing the figures \ref{waitTime_vs_beta1}-\ref{usedCars_vs_AR}  together, we note that our approach reduce the number of active vehicles significantly as compared to the other baselines, and achieves the smallest waiting times. The additional penalty in travel time is small as compared to the approaches which do not allow ride-share. As more users are accepted, the proposed policy uses more vehicles, and the travel and idle times increase as expected. Since more vehicles are used and the idle time being larger indicates that the vehicles are finding better requests to merge, the waiting times per request decrease.

 The effect of time overhead ($\delta_t$) on waiting time, idle time and reject rate 
 is captured in Figures \ref{waitTime_vs_overhead}--\ref{rr_vs_overhead}, which are obtained by changing the values of $\beta_3$. Note that both NoDS-NoRS and DS-NoRS policies are not plotted since carpooling is not allowed, and hence $\delta_t=0$. The  extra travel time ($\delta_t$) is due to participating in ride-sharing. Intuitively, as $\delta_t$ increases, the waiting time decreases as more customers can be served using a single car and by better utilizing the users proximity. However, this comes at an additional increased in the idle time of vehicles. By allowing larger travel time overhead, cars behave greedily and pool more customers which results in increasing the idle times of other vehicles as depicted in Figure \ref{idleTime_vs_overhead}.    
%
%
%
On the other hand, by enforcing the overhead to be small, DeepPool will likely dispatch the vehicles to their destination without changing the route throughout the trip. Hence, customers may wait longer in order to be served by some other vehicles. Moreover, when the extra time increases, more customers can be served and thus the reject rate is expected to decrease. The gain of the DeepPool over the NoDS-RS policy is remarkable for all values of $\delta_t$ (see Figure \ref{rr_vs_overhead}) which highlights the benefits of ride-sharing in improving customer satisfaction at small travel time overhead.   

\textcolor{black}{
Figures \ref{waitTime_vs_waiting}-\ref{idleTime_vs_days} compare the wait time and
idle time for different policies for one day (Sunday) from $6$pm till $6$am next day.
We see that DeepPool consistently performs the best (close to the actual-demand) compared to other algorithms. This implies that our policy learns the optimal dispatch policy and thus can manage dispatching vehicles for improved ride-sharing services.
\\
Figures \ref{Utili_mean}-\ref{Utili_min} show the mean and minimum of the vehicle utilization
rate (i.e., percentage of time at which
a vehicle is busy serving customers). We used  $6000$ vehicles. We see that DeepPool achieves better utilization, for both mean and minimum performance cases, compared to other approaches. Further, the gap between the actual demand and DeepPool (in both scenarios) is small. This performance emphasizes that our approach learns the optimal policy for an individual
vehicle, thus every vehicle aims to take the best action to maximize its own reward.
Thus, DeepPool tries to fairly distribute the rewards across the
different vehicles.
}


\subsection{Discussion}

The preceding results show the advantages of DeepPool: a distributed model-free DQN-based approach. 
Since ride-sharing platforms like Uber and Lyft give the flexibility to drives to choose where to go, DeepPool is more realistic (compared to centralized and traditional methods) as each vehicle can take the desired action to maximize its own revenue, given its current state and thus more practical to be implemented. Further, since DeepPool is not centralized, it takes much less time compared to centralized based approaches \cite{oda2018movi} when taking an action.
Moreover, centralized approaches may not result in choosing the optimal
action for each one individually, although it may better optimize the whole system. 
Additionally, DeepPool allows to share cost of traveling among passengers, e.g., fuel and highway tolls. It also reduces the traffic congestion by better use of vehicles seats.

DeepPool can be extended to cover the scenarios where there are  multiple classes of users. For example, we can adapt $\beta_3$ to be user specific in the reward (cf. (\ref{reward_veh_n})) if a user does not want to spend extra time for carpooling. Also, $\beta_3$ can also be controlled externally by giving incentives to the users to opt for the carpooling. 
Finally, our model can also be extended to accommodate the scenarios where $\beta_i$, $ i=1,\ldots,4$ is different for each vehicle $n$ in order to capture the different behaviors of the drivers and their preferences. However, the complete analysis is left for the future.

\section{Conclusions}

We propose DeepPool as a distributed model-free framework, which uses deep neural networks and reinforcement learning to learn optimal dispatch policies by interacting with the external environment. It efficiently incorporates travel demand statistics and deep learning models to manage dispatching vehicles for improved ride sharing services. Using real-world dataset of taxi trip records in New York City, DeepPool performs better than other strategies that do not consider ride sharing or dispatching the vehicles to locations where the predicted future demand is more likely. Further, DeepPool is distributed and thus can adapt rapidly to dynamic environment. As a future work, interaction among multiple service providers is an interesting future direction.  The dispatch decision where the vehicles serve each passenger in a multi-hop (where passengers are dropped multiple times before reaching there destination) constitutes an important future research direction. The characterization of an optimal incentive mechanism to influence users to opt for carpooling is another interesting future research direction.  

\bibliographystyle{IEEEtran}
\bibliography{bib}
	
\end{document}